\documentclass[sigconf]{acmart}
\usepackage{microtype}
\usepackage{graphicx}
\usepackage{subfigure}
\usepackage{booktabs} 
\usepackage{algorithm}
\usepackage{algorithmic}
\usepackage{tabularx}
\usepackage{adjustbox}
\usepackage{hyperref}

\usepackage{amsmath}
\usepackage{mathtools}
\usepackage{amsthm}
\usepackage{amsfonts}       
\usepackage[utf8]{inputenc} 
\usepackage[T1]{fontenc}    
\usepackage{url}            
\usepackage{nicefrac}       
\usepackage{xcolor}         
\usepackage{bbm}
\usepackage[utf8]{inputenc}
\usepackage[T1]{fontenc}    
\usepackage{xcolor}         
\usepackage{natbib}
\usepackage{xspace}
\usepackage{enumerate, paralist}
\usepackage{natbib}
\usepackage{comment}
\usepackage{multirow}
\usepackage{xspace}
\usepackage{rotating}
\usepackage{booktabs} 
\usepackage{tabularx}
\usepackage{enumitem}
\usepackage{thmtools} 
\usepackage{thm-restate}
\theoremstyle{plain}

\newtheorem{theorem}{Theorem}[section]
\newtheorem{condition}{Condition}[section]
\newtheorem{remark}{Remark}[section]
\newtheorem{lemma}[theorem]{Lemma}

\newtheorem{assumption}[condition]{Assumption}
\newtheorem{corollary}[theorem]{Corollary}

\theoremstyle{definition} 
\newtheorem{definition}[theorem]{Definition}

\newcommand{\yunzhe}[1]{{\textsf{\textcolor{purple}{[From Yunzhe: #1]}}}}

\newcommand{\bx}{\mathbf{x}}
\newcommand{\bh}{\mathbf{h}}
\newcommand{\BX}{\mathbf{X}}
\newcommand{\bw}{\mathbf{W}}
\newcommand{\bD}{\mathbf{D}}

\newcommand{\hi}{\widehat{i}}

\newcommand{\bbr}{\mathbb{R}}
\newcommand{\bbe}{\mathbb{E}}

\newcommand{\caln}{\mathcal{N}}
\newcommand{\para}[1]{\noindent \textbf{#1}\xspace}

\newcommand{\calo}{\mathcal{O}}
\newcommand{\calt}{\mathcal{T}}
\newcommand{\call}{\mathcal{L}}

\newcommand{\wcalo}{\widetilde{\calo}}
\newcommand{\wcaln}{\widehat{\caln}}
\newcommand{\bbh}{\mathbf{H}}

\newcommand{\btheta}{\theta}
\newcommand{\htheta}{\widehat{\theta}}

\newcommand{\deter}{\text{\normalfont det}}

\newcommand{\hd}{\widetilde{d}}

\newcommand{\bsg}{\mathbf{G}}

\newcommand{\bsr}{\mathbf{r}}

\newcommand{\hx}{h(\mathbf{x}_t)}

\newcommand{\gx}{g(\mathbf{x}_t; \btheta_0)}

\newcommand{\sysn}{\text{M-CNB}\xspace}


\copyrightyear{2024}
\acmYear{2024}
\setcopyright{rightsretained}
\acmConference[KDD '24]{Proceedings of the 30th ACM SIGKDD Conference on Knowledge Discovery and Data Mining}{August 25--29, 2024}{Barcelona, Spain}
\acmBooktitle{Proceedings of the 30th ACM SIGKDD Conference on Knowledge Discovery and Data Mining (KDD '24), August 25--29, 2024, Barcelona, Spain}
\acmDOI{10.1145/3637528.3671691}
\acmISBN{979-8-4007-0490-1/24/08}

\makeatletter
\gdef\@copyrightpermission{
  \begin{minipage}{0.3\columnwidth}
   \href{https://urldefense.com/v3/__https://creativecommons.org/licenses/by-nc-sa/4.0/}{\includegraphics[width=0.90\textwidth]{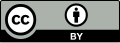}}
  \end{minipage}\hfill
  \begin{minipage}{0.7\columnwidth}
   \href{https://urldefense.com/v3/__https://creativecommons.org/licenses/by-nc-sa/4.0/}{This work is licensed under a Creative Commons Attribution International 4.0 License.}
  \end{minipage}
  \vspace{5pt}
}
\makeatother

\begin{document}

\title{Meta Clustering of Neural Bandits}

\settopmatter{authorsperrow=3}

\author{Yikun Ban}
\authornote{Both authors contributed equally to this paper.}
\email{yikunb2@illinois.edu}
\affiliation{%
  \institution{University of Illinois at Urbana-Champaign}
   \city{Urbana}
     \state{IL}
  \country{USA}
}

\author{Yunzhe Qi}
\authornotemark[1]
\email{yunzheq2@illinois.edu}
\affiliation{%
  \institution{University of Illinois at Urbana-Champaign}
   \city{Urbana}
     \state{IL}
  \country{USA}
}

\author{Tianxin Wei}
\email{twei10@illinois.edu}
\affiliation{%
  \institution{University of Illinois at Urbana-Champaign}
   \city{Urbana}
     \state{IL}
  \country{USA}
}

\author{Lihui Liu}
\email{lihuil2@illinois.edu}
\affiliation{%
  \institution{University of Illinois at Urbana-Champaign}
   \city{Urbana}
     \state{IL}
  \country{USA}
}

\author{Jingrui He}
\email{jingrui@illinois.edu}
\affiliation{%
  \institution{University of Illinois at Urbana-Champaign}
   \city{Urbana}
     \state{IL}
  \country{USA}
}


\begin{abstract}
The contextual bandit has been identified as a powerful framework to formulate the recommendation process as a sequential decision-making process, where each item is regarded as an arm and the objective is to minimize the regret of $T$ rounds.
In this paper, we study a new problem, 
Clustering of Neural Bandits, by extending previous work to the arbitrary reward function, to strike a balance between user heterogeneity and user correlations in the recommender system. To solve this problem, we propose a novel algorithm called M-CNB, which utilizes a meta-learner to represent and rapidly adapt to dynamic clusters, along with an informative Upper Confidence Bound (UCB)-based exploration strategy.  
We provide an instance-dependent performance guarantee for the proposed algorithm that withstands the adversarial context, and we further prove the guarantee is at least as good as state-of-the-art (SOTA) approaches under the same assumptions. 
In extensive experiments conducted in both recommendation and online classification scenarios, M-CNB outperforms SOTA baselines. This shows the effectiveness of the proposed approach in improving online recommendation and online classification performance.
\end{abstract}


\maketitle

\section{Introduction}

Recommender systems play an integral role in various online businesses, including e-commerce platforms and online streaming services. They leverage user correlations to assist the perception of user preferences, a field of study spanning several decades. In the past, considerable effort has been directed toward supervised-learning-based collaborative filtering methods within relatively static environments \citep{su2009survey, he2017neural}.
However, the ideal recommender systems should adapt over time to consistently meet user interests. Consequently, it is natural to formulate the recommendation process as a sequential decision-making process. In this paradigm, the recommender engages with users, observes their online feedback (i.e., rewards), and optimizes the user experience for long-term benefits, rather than fitting a model on the collected static data based on supervised learning \cite{xue2023resact,chen2022off,gao2023alleviating}.
Based on this idea,
this paper focuses on the formulation of contextual bandits, where each item is treated as an arm (context) in a recommendation round, and the primary objective is to minimize the cumulative regret over $T$ rounds and tackle the dilemma of exploitation and exploration in the sequential
decision-making process
\cite{ban2023neural,2011improved,2016collaborative,2014onlinecluster,2016collaborative,gentile2017context,2019improved,ban2021local, qi2022neural,mcdonald2023impatient,qi2023graph,ban2020generic}.


Linear contextual bandits model a user's preference through a linear reward function based on arm contexts \cite{dani2008stochastic, 2010contextual, 2011improved}. However, given the substantial growth of users in recommender systems, it can be overly ambitious to represent all user preferences with a single reward function, and it may overlook the user correlations if each user is modeled as a single bandit.
To address this challenge, a series of methods known as clustering of linear bandits \citep{2014onlinecluster,2016collaborative,gentile2017context,2019improved,ban2021local} have emerged, which represent each cluster of users as a reward function, achieving a balance between user heterogeneity and user correlations. Note that the cluster information is unknown in this problem setting.
In essence, with each user being treated as a linear contextual bandit, these methods adopt graph-based techniques to dynamically cluster users, and leverage user correlations for making arm recommendations.
However, it is crucial to acknowledge the limitations of this line of works: they all rely on linear reward functions, and user clusters are represented as linear combinations of individual bandit parameters. 
The assumptions of linearity in reward functions and the linear representation of clusters may not hold up well in real-world applications \citep{valko2013finite,zhou2020neural}.

In relaxation of the assumption on reward mapping functions,
inspired by recent advances in the single neural bandit \cite{zhou2020neural, zhang2020neural} where a neural network is assigned to learn \textit{an} unknown reward function, we study the new problem of Clustering of Neural Bandits (CNB) in this paper.
Different from the single neural bandit \cite{zhou2020neural, zhang2020neural} and clustering of linear bandits \citep{2014onlinecluster,2016collaborative,gentile2017context,2019improved,ban2021local}, CNB introduces the bandit clusters built upon the \textit{arbitrary} reward functions, which can be either linear or non-linear.
Meanwhile, we note that the underlying clusters are usually not static over specific arm contexts \citep{2016collaborative}.
For example, in the personalized recommendation task, two users (bandits) may both like "country music", but can have different opinions on "rock music". Therefore,
adapting to arm-specific "relative clusters" in a dynamic environment is one of the main challenges in this problem. We propose a novel algorithm, Meta Clustering of Neural Bandits (\sysn), to solve the CNB problem.
Next, we will summarize our key ideas and contributions.

\textbf{Methodology}. 
To address the CNB problem, we must confront three key challenges:
(1) Efficiently determining a user's relative group:
Our approach involves employing a neural network, named the "user learner," to estimate each user's preferences. By grouping users with similar preferences, we efficiently create clusters with a process taking $\calo(n)$ time, where $n$ is the number of bandits (users).
(2) Effective parametric representation of dynamic clusters:
Inspired by advancements in meta-learning \citep{finn2017model, yao2019hierarchically}, we introduce a meta-learner capable of representing and swiftly adapting to evolving clusters. In each round $t$, the meta-learner leverages its perceived knowledge from prior rounds $\{1, \dots, t-1\}$ to rapidly adapt to new clusters via a few samples. This enables the rapid acquisition of nonlinear cluster representations, marking our first main contribution.
(3) Balancing exploitation and exploration with relative bandit clusters:
Our second main contribution is proposing an informative UCB-type exploration strategy, which takes into account both user-side and meta-side information for balancing the exploration and exploitation.
By addressing these three main challenges, our approach manages to solve the CNB problem effectively and efficiently.

\textbf{Theoretical analysis}.
To obtain a regret upper bound for the proposed algorithm, we need to tackle the following three challenges:
(1) Analyzing neural meta-learner in bandit framework: To finish the analysis, we must build a confidence ellipsoid for the meta-learner approximation, which is one of the main research gaps. To deal with this gap, we bridge the meta-learner and user-learner via the Neural Tangent Kernel (NTK) regression and build the confidence ellipsoid upon the user-learner, which allows us to achieve a more comprehensive understanding of the meta-learner's behavior.
(2) Reducing the naive $ \wcalo(\sqrt{nT})$ regret upper bound: $\wcalo(\sqrt{T})$ is roughly the regret effort to learn a single neural bandit, and thus $\wcalo(\sqrt{nT})$ are the regret efforts to learn $n$ neural bandits for $n$ users. We reduce the  $\wcalo(\sqrt{nT})$ efforts to $\wcalo(\sqrt{qT})$, where $q$ is the expected number of clusters. This also indicates the proposed algorithm can leverage the collaborative effects among users.
(3) Adversarial attack on contexts:
In most neural bandit works, a common assumption is that the NTK matrix is non-singular, requiring that no two observed contexts (items) are identical or parallel \cite{zhou2020neural, zhang2020neural}. This vulnerability makes their regret analysis susceptible to adversarial attacks and less practical in real-world scenarios. 
In face of this challenge, we provide an instance-dependent regret analysis that withstands the context attack, and allows the contexts to be repeatedly observed.  Furthermore, under the same assumptions as in existing works, we demonstrate that our regret upper bound is at least as good as SOTA approaches.
The above efforts to address the challenges in the theoretical analysis is our third main contribution.

\textbf{Evaluations}.
We evaluate the proposed algorithm in two scenarios: Online recommendation and Online classification with bandit feedback.
For the first scenario, which naturally lends itself to CNB, we assess the algorithm's performance on four recommendation datasets. 
Since online classification has been widely used to evaluate neural bandits \cite{zhou2020neural,zhang2020neural,ban2021ee}, we evaluate the algorithms on eight classification datasets where each class can be considered as a bandit (user), 
and correlations among classes are expected to be exploited. We compare the proposed algorithm with 8 strong baselines and show the superior performance of the proposed algorithm. 
Additionally, we offer the empirical analysis of the algorithm's time complexity, and conduct extensive sensitivity studies to investigate the impact of critical hyperparameters. 
The above empirical evaluation is our fourth main contribution.

Next, detailed discussion regarding related works is placed in Section \ref{sec:related}.
After introducing the problem definition in Section \ref{sec:prod}, we present the proposed algorithm, \sysn, in Section \ref{sec:algorithm} together with theoretical analysis in Section \ref{sec:theo1}. Then, we provide the experimental results in Section \ref{sec:exp} and conclude the paper in Section \ref{sec:conclusion}. 

\section{ Related Work} \label{sec:related}

In this section, we briefly review the related works, including clustering of bandits and neural bandits.

\textbf{Clustering of bandits}. CLUB~\cite{2014onlinecluster} first studies collaborative effects among users in contextual bandits where each user hosts an unknown vector to represent the behavior based on the linear reward function. CLUB formulates user similarity on an evolving graph and selects an arm leveraging the clustered groups. Then, \citet{2016collaborative,gentile2017context} propose to cluster users based on specific contents and select arms leveraging the aggregated information of conditioned groups. \citet{2019improved} improves the clustering procedure by allowing groups to split and merge. \citet{ban2021local} uses seed-based local clustering to find overlapping groups, different from global clustering on graphs. \citet{korda2016distributed,yang2020exploring,wu2021clustering, liu2022federated, wang2024online} also study clustering of bandits with various settings in recommender systems. However, all these works are based on the linear reward assumption, which may fail in many real-world applications. 

\textbf{Neural bandits}. 
\citet{lipton2018bbq, riquelme2018deep} adapt the Thompson Sampling (TS) to the last layer of deep neural networks to select an action. However, these approaches do not provide regret analysis. \citet{zhou2020neural} and  \citet{zhang2020neural} first provide the regret analysis of UCB-based and TS-based neural bandits, where they apply ridge regression on the space of gradients.
\citet{ban2021multi} studies a multi-facet bandit problem with a UCB-based exploration.
\citet{jia2021learning} perturbs the training samples for incorporating both exploitation and exploration.
EE-Net \citep{ban2021ee,ban2023neural2} proposes to use another neural network for exploration with applications on active learning \cite{ban2022improved, ban2024neural} and meta-learning \cite{qi2023meta}.
\cite{xu2020neural} combines the last-layer neural network embedding with linear UCB to improve the computation efficiency.
\citet{dutta2019automl} uses an off-the-shelf meta-learning approach to solve the contextual bandit problem in which the expected reward is formulated as Q-function. 
\citet{santana2020contextual} proposes a Hierarchical Reinforcement Learning framework for recommendation in the dynamic experiments, where a meta-bandit is used for the selected independent recommender system.
\citet{kassraie2022neural} revisit Neural-UCB type algorithms and shows the $\widetilde{\calo}(\sqrt{T})$ regret bound without the  restrictive
assumptions on the context. 
\citet{maillard2014latent, hong2020latent} study the latent bandit problem where the reward distribution of arms are conditioned on some unknown discrete latent state and prove the $\widetilde{\calo}(\sqrt{T})$  regret bound for their algorithm as well.
Federated bandits \citep{dai2022federated} consider dealing with multiple bandits (agents) while preserving the privacy of each bandit.
\citet{deb2023contextual} reduce the contextual bandits to neural online regression for tighter regret upper bound.
\citet{qi2023graph} propose to use graph to formulate user correlations with the adoption of graph neural networks. 
However, the above works either focus on the different problem settings or overlook the clustering of bandits.

\textbf{Other related works}. 
\citep{kveton2021meta, simchowitz2021bayesian} study meta-learning in Thompson sampling and \citet{hong2022hierarchical, wan2021metadata} aims to exploit the hierarchical knowledge among hierarchical Bayesian bandits. However, they focus on the Bayesian or non-contextual bandits.

\section{Problem: Clustering of Neural Bandits} \label{sec:prod}

In this section, we introduce the CNB problem, motivated by learning correlations among bandits with arbitrary reward functions. Next, we will use the scenarios of personalized recommendation to state the problem setting.

Suppose there are $n$ users (bandits), $N = \{1, \dots, n\}$, to serve on a platform. In the $t^{\textrm{th}}$ round, the platform receives a user $u_t \in N$ (unique ID for this user) and prepares the corresponding $K$ candidate arms $\BX_t = \{ \bx_{t, 1}, \bx_{t, 2}, \dots, \bx_{t,K} \}$. Each arm is represented by its $d$-dimensional feature vector  $\bx_{t, i} \in \bbr^d,  i \in [K] = \{1, \dots, K\}$, which will encode the information from both the user side and the arm side \cite{2010contextual}. 
Then, the learner is expected to select an arm $\bx_{t} \in \BX_t$ and recommend it to $u_t$, where $u_t$ refers to the target or served user. In response to this action, $u_t$ will provide the platform with a corresponding reward (feedback) $r_{t}$. Here, since different users may generate different rewards towards the same arm, we use $r_{t,i}| u_t$ to represent the reward produced by $u_t$ given $\bx_{t,i}$. The formal definition of arm reward is below.

Given $u_t \in N$, the reward $r_{t,i}$ for each candidate arm $\bx_{t,i} \in \BX_t$ is assumed to be governed by an unknown function by
\begin{equation} 
 r_{t,i}|u_t =   h_{u_t}(\bx_{t,i}) + \zeta_{t, i},
 \end{equation}
where $h_{u_t}$ is an unknown reward function associated with $u_t$, and it can be either linear or non-linear. $\zeta_{t, i}$ is a noise term with zero expectation $\bbe[\zeta_{t, i}] = 0$. We also assume the reward $r_{t,i} \in [0, 1]$ is bounded, as in many existing works \citep{2014onlinecluster,gentile2017context,ban2021local}. Note that previous works on clustering of linear bandits all assume $h_{u_t}$ is a linear function with respect to arm $\bx_{t,i}$ \citep{2014onlinecluster, 2016collaborative, gentile2017context, 2019improved, ban2021local}.

Meanwhile, users may exhibit clustering behavior. 
Inspired by \citep{gentile2017context, 2016collaborative}, we consider the cluster behavior to be item-varying, i.e., the users who have the same preference on a certain item may have different opinions on another item. Therefore, we formulate a set of users with the same opinions on a certain item as a relative cluster, with the following definition.
\begin{definition}[Relative Cluster]  \label{def:group}
In round $t$, given an arm $\bx_{t,i} \in \BX_t$, a relative cluster $\caln(\bx_{t,i}) \subseteq N$ with respect to $\bx_{t,i}$ satisfies
\[
\begin{aligned}
    &(1) \ \forall u, u' \in \caln(\bx_{t,i}),   \bbe[r_{t,i}| u] = \bbe[r_{t,i}| u']\\
    &(2) \  \nexists \ \caln' \subseteq N,  \text{s.t.}   \ \caln' \ \text{satisfies} \ (1) \ \text{and} \  \caln(\bx_{t,i}) \subset \caln'.
\end{aligned}
\]
\end{definition}
The condition $(2)$ is to guarantee that no other clusters contains $\caln(\bx_{t,i})$.
This cluster definition allows users to agree on certain items while disagree on others, which is consistent with the real-world scenario. Since the users from different clusters are expected to have distinct behavior with respect to $\bx_{t,i}$, we provide the following constraint among relative clusters.

\begin{definition} [$\gamma$-gap]  \label{def:gap}
Given two different cluster $\caln(\bx_{t,i})$, $\caln'(\bx_{t,i})$, there exists a constant $\gamma > 0$, such that
\[
\forall u \in \caln(\bx_{t,i}), u' \in \caln'(\bx_{t,i}),   |  \bbe[r_{t,i}| u] -   \bbe[r_{t,i}| u'] | \geq \gamma.
\]
\end{definition}

For any two clusters in $N$, we assume that they satisfy the $\gamma$-gap constraint. Note that such an assumption is standard in the literature of online clustering of bandit to differentiate clusters \citep{2014onlinecluster, 2016collaborative, gentile2017context, 2019improved, ban2021local}. 
As a result, given an arm $\bx_{t,i}$, the bandit pool $N$ can be divided into $q_{t,i}$ \textit{non-overlapping} clusters: $\caln_1(\bx_{t,i}), \caln_2(\bx_{t,i}),$ $\dots, \caln_{q_{t,i}}(\bx_{t,i})$, where $q_{t,i} \ll  n$. 
Note that the cluster information is \textit{unknown} in the platform.

For the CNB problem, the goal of the learner is to minimize the pseudo regret of $T$ rounds:
\begin{equation} \label{eq:regret}
\mathbf{R}_T = \sum_{t=1}^T \bbe[ r_t^\ast  - r_t \mid u_t, \BX_t],  
\end{equation}
where $r_t$ is the reward received in round $t$, and $\bbe[r_t^\ast |u_t, \BX_t] = \max_{\bx_{t, i} \in \BX_t} h_{u_t}(\bx_{t,i})$.

\textbf{Notations}. 
Let $\bx_t$ be the arm selected in round $t$, and $r_t$ be the corresponding reward received in round $t$.
We use $\|\bx_t\|_2$  to represent the Euclidean norm.
For each user $u \in N$, let $\mu_t^u$ be the number of rounds that user $u$' learner has been served up to round $t$, and $\calt^u_{t}$ be all of $u$'s historical data up to round $t$.
$m$ is the width of neural network and $L$ is depth of neural network in the proposed approach.
Given a group $\caln$, all its data up to round $t$ can be denoted by $ \{\calt^u_t \}_{u \in \caln}  =  \{ \calt_t^u | u \in \caln \}$. We use standard $\mathcal{O}$ and $\Omega$ notation to hide constants.


\section{Proposed Algorithm} \label{sec:algorithm}

In this section, we present our proposed algorithm, denoted as \sysn, to address the formulated CNB problem. \sysn leverages the potential correlations among bandits, and aims to rapidly acquire a representation for dynamic relative clusters.

For \sysn, we utilize a meta-learner, denoted as $\Theta$, to rapidly adapt to clusters, as well as represent the behavior of a cluster. Additionally, there are $n$ user-learners, denoted by $\{\theta^u\}_{u \in N}$, responsible for learning the preference $h_u(\cdot)$ for each user $u \in N$. 
In terms of the workflow, 
the primary role of the meta-learner is to determine recommended arms, while the user-learners are primarily utilized for clustering purposes. 
The meta-learner and user-learners share the same neural network structure, denoted as $f$.
And the workflow of \sysn is divided into three main components: User clustering, Meta adaptation, and UCB-based selection. Then, we proceed to elaborate their details.


\begin{figure}[!t] 
    \centering
    \includegraphics[height=40mm, width=70mm]{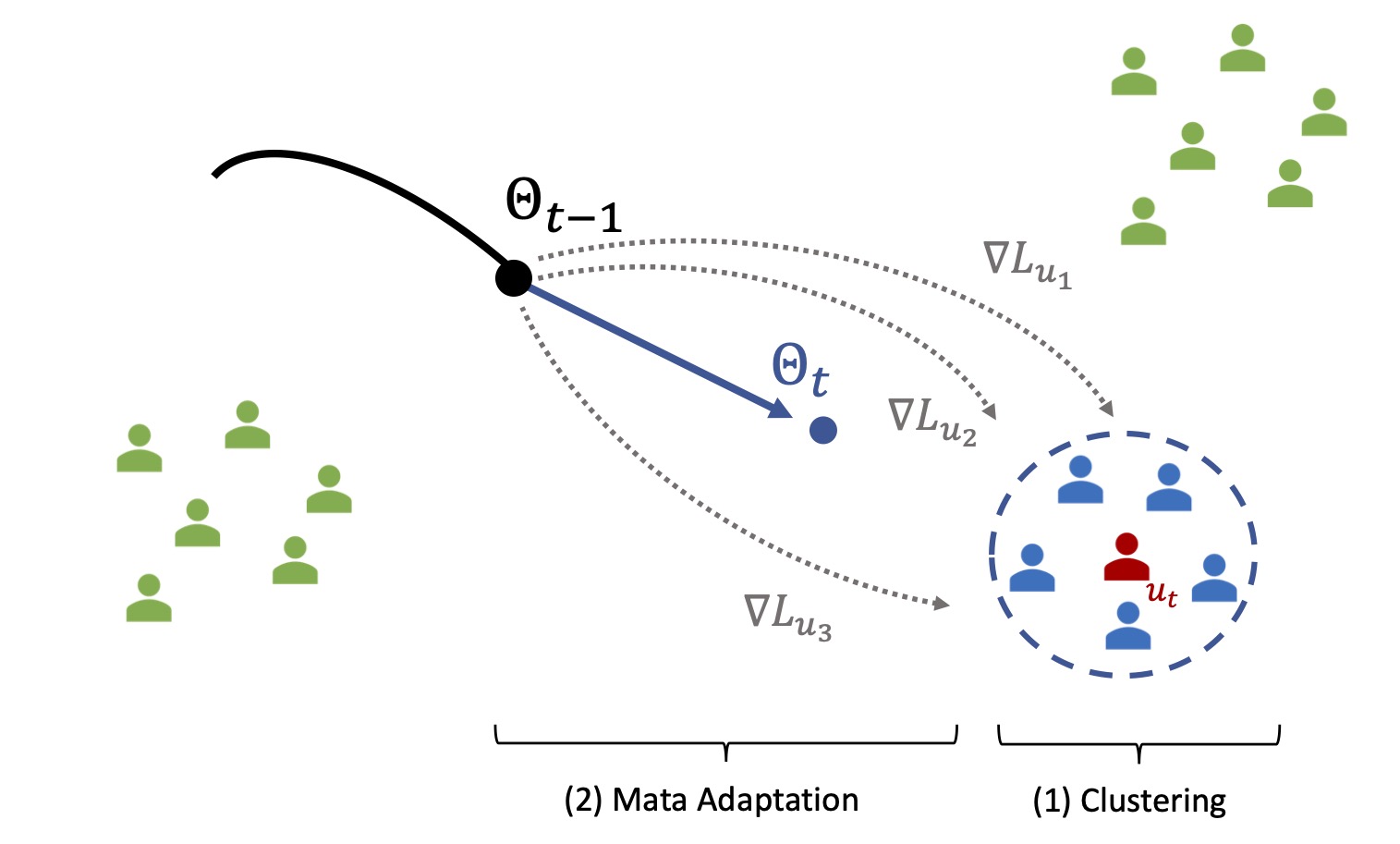}
     \vspace{-1em}
    \caption{ Clustering and Meta Adaptation: Given $u_t$ and an arm $\bx_{t,i}$, (1) \sysn identifies cluster $\wcaln_{u_t}(\bx_{t,i})$, and then (2) meta-learner $\Theta_{t-1}$ rapidly adapt to this cluster, proceeding to (3) the UCB exploration. }
       \label{fig:workflow}
\end{figure}

\textbf{User clustering}.  
Recall that in Section \ref{sec:prod}, each user $u \in N$ is governed by an unknown function $h_u$. In this case, we use a neural network $f(\cdot; \theta^{u})$, to estimate $h_u$. 
In round $t \in [T]$, let $u_t$ be the user to serve. Given $u_t$'s past data up to round $t-1$, i.e., $\calt_{t-1}^{u_t}$, we can train parameters $\theta^{u_t}$  by minimizing the following loss: 
$
\call \left(\theta^{u_t}  \right) =  \sum_{(\bx, r) \in  \calt_{t-1}^{u_t} }  ( f(\bx; \theta^{u_t}) - r)^2/2. 
$
Let $\theta_{t-1}^{u_t}$ represent $\theta^{u_t}$ trained on $\calt_{t-1}^{u_t}$ in round $t-1$ by stochastic gradient descent (SGD).
Therefore, for each $u \in N$, we can obtain the trained parameters $\theta^u_{t-1}$. Then,
given $u_t$ and an arm $\bx_{t,i}$, we return $u_t$'s estimated cluster with respect to arm $\bx_{t,i}$ by 
\begin{equation} \label{eq:group}
\begin{aligned}
&\widehat{\caln}_{u_t}(\bx_{t,i}) = \big\{ u \in N \ \big| \ |f(\bx_{t,i}; \theta^u_{t-1}) -  f(\bx_{t,i}; \theta^{u_t}_{t-1})| \leq \frac{\nu-1}{\nu} \gamma \big\}.
\end{aligned}
\end{equation}
where $\gamma \in (0, 1)$ represents the assumed $\gamma$-gap and $\nu > 1$ is a tuning parameter to for the exploration of cluster members.

\begin{algorithm}[t]
\begin{algorithmic}[1] 
\caption{ \sysn }\label{alg:main}
\STATE {\bfseries Input:}  $T$ (number of rounds), $\gamma, \nu$ (cluster exploration parameter),  $ S$ (norm parameter),  $\delta$ (confidence level) ,  $\eta_1, \eta_2$ (learning rate), $m$(width of neural network). 
\STATE Initialize $\Theta_0$;     $\theta_0^u =\Theta_0, \mu_0^{u} = 0, \mathcal{T}^{u}_{0} = \emptyset, \forall u \in N$ 
\STATE Observe one data for each $ u \in N$ 
\FOR{  $t = 1, 2, \dots, T$}
\STATE Receive a target user $u_t \in N$ and observe $k$ arms $\BX_t =\{\bx_{t, 1}, \dots, \bx_{t, k}\}$ 
\FOR{ $i \in [k]$ }
\STATE Determine $\widehat{\caln}_{u_t}(\bx_{t,i}) =  \{ u \in N \ | \ |f(\bx_{t,i};  \theta^u_{t-1}) -  f(\bx_{t,i}; \theta^{u_t}_{t-1})| \leq \frac{\nu-1}{\nu} \gamma \}$. 
\STATE  $ \Theta_{t,i}  =  \text{SGD\_Meta} \left(\widehat{\caln}_{u_t}(\bx_{t,i}),  \Theta_{t-1} \right)   $  
\STATE  $\mathbf{U}_{t, i} = f(\bx_{t,i}; \Theta_{t,i})  +  \frac{   \| \nabla_{\Theta} f (\bx_{t,i}; \Theta_{t, i}) - \nabla_{\theta} f (\bx_{t,i}; \theta_{0}^{u_t})\|_2 }{m^{1/4}}  +  \sqrt{ \frac{S  + 1 } {2\mu^{u}_{t}}} +   \sqrt{\frac{  2 \log(  1 /\delta) }{\mu^{u}_t}}   $ 
\ENDFOR
\STATE $ \hi = \arg_{ i \in [k]} \max  \mathbf{U}_{t, i}$ 
\STATE Play $\bx_{t, \hi}$ and observe reward $r_{t, \hi}$ 
\STATE  $\bx_t =   \bx_{t, \hi},  \ r_t  =  r_{t, \hi},  \   \Theta_t = \text{SGD\_Meta} \left(\widehat{\caln}_{u_t}(\bx_{t,\hi}),  \Theta_{t-1} \right) $ 
\vspace{0.5em}
\FOR{$u \in \widehat{\caln}_{u_t}(\bx_{t})$}
\STATE  $\call_t \left(\theta_t^{u} \right) =  ( f(\bx_t; \theta_t^u) - r_t)^2/2  $ 
\STATE $\theta^{u}_t =  \theta_t^u -  \eta_1 \triangledown_{\theta_t^u} \call_t \left(\theta_t^u  \right)$    \ \ \# User Adaptation
\STATE $\mu_t^{u} = \mu_{t-1}^{u} + 1$ ,  $\mathcal{T}^{u}_{t} =  \mathcal{T}^{u}_{t-1} \cup \{ (\bx_t, r_t)\} $
\ENDFOR
\vspace{0.5em}
\FOR{ $u \not \in \widehat{\caln}_{u_t}(\bx_{t})$ }
\STATE  $\theta^{u}_t =\theta^{u}_{t-1}$, \ $\mu_t^{u} = \mu_{t-1}^{u}$ ,  $\mathcal{T}^{u}_{t} =  \mathcal{T}^{u}_{t-1}$
\ENDFOR
\ENDFOR
\end{algorithmic}
\end{algorithm}

\textbf{Meta adaptation}. We employ one meta-learner $\Theta$ to represent and adapt to the behavior of dynamic clusters.
In meta-learning, the meta-learner is trained based on a number of different tasks and can quickly adapt to new tasks with a small amount of new data \citep{finn2017model}.
Here, we consider a cluster  $\caln_{u_t}(\bx_{t,i})$ as a task and its collected data as the task distribution. 
As a result, \sysn has two adaptation phases: meta adaptation, and user adaptation.

\textit{Meta adaptation}. In the $t^{\textrm{th}}$ round, given a cluster $\widehat{\caln}_{u_t}(\bx_{t,i})$,
we have the available "task distributions" $\{ \calt_{t-1}^u \}_{u \in \widehat{\caln}_{u_t}(\bx_{t,i})}$.
The goal of the meta-learner is to quickly adapt to the bandit cluster. 
Thus, we randomly draw a few samples from   $\{ \calt_{t-1}^u \}_{u \in \widehat{\caln}_{u_t}(\bx_{t,i})}$ and update $\Theta$ in round $t$ using SGD, denoted by $\Theta_{t, i}$, based on $\Theta_{t-1}$ that is continuously trained on the collected interactions to incorporate the knowledge of past $t-1$ rounds.
The workflow is described in Figure \ref{fig:workflow} and Algorithm \ref{alg:meta}.

\textit{User adaptation}. In the $t^{\textrm{th}}$ round, given $u_t$,  after receiving the reward $r_t$, we have available data $(\bx_t, r_t)$. Then, the user leaner $\theta^{u_t}$ is updated in round $t$ to have a refined clustering capability, 
denoted by $\theta^{u_t}_{t}$.
As the users in a cluster share the same or similar preferences on a certain item, we update all the user learners in this cluster, described in Algorithm \ref{alg:main} Lines 14-18.

Note that for the clustering of linear bandits works \citep{2014onlinecluster,2016collaborative,gentile2017context,2019improved,ban2021local}, they represent the cluster behavior $\Theta$ by the linear combination of bandit-learners, e.g., $ \Theta = \frac{1}{ | \widehat{\caln}_{u_t}(\bx_{t,i})|} \sum_{u \in \widehat{\caln}_{u_t}(\bx_{t,i})} \theta^u_{t}$. This can lead to limited representation power of the cluster learner, and their linear reward assumptions may not necessarily hold for real world settings \cite{zhou2020neural}. Instead, we use the meta adaptation to update the meta-learner $\Theta_{t-1}$ according to $\widehat{\caln}_{u_t}(\bx_{t,i})$, which can represent non-linear combinations of user-learners \citep{finn2017model, wang2020onglobal}.


\begin{algorithm}[t]
\begin{algorithmic}
\caption{SGD\_Meta ($\widehat{\caln}_{u_t}(\bx_{t,i}), \Theta_{t-1}$) }\label{alg:meta}
\STATE $ \widehat{\caln} = \widehat{\caln}_{u_t}(\bx_{t,i})$
\FOR{ $u \in \widehat{\caln}$}
\STATE Randomly draw $(\bx^u, r^u)$ from $\mathcal{T}^{u}_{t-1}$ \\
\STATE $\call_u \left(\Theta_{t-1} \right) =  ( f(\bx^u; \Theta_{t-1}) - r^u)^2/2$\\
\ENDFOR
\STATE $\call_{t-1}(\widehat{\caln})= \frac{1}{| \widehat{\caln}|}  \sum_{u \in \widehat{\caln}} \call_u \left( \Theta_{t-1}  \right)$   \\ 
\STATE $\Theta_{t, i} = \Theta_{t-1} -  \eta_2 \nabla_{ \Theta_{t-1} }\call_{t-1}(\widehat{\caln})$   \ \ \# Meta Adaptation           \\
 \textbf{Return:} $\Theta_{t,i}$ \\
 \end{algorithmic}
\end{algorithm}

\textbf{UCB-based Exploration}.
To balance the trade-off between the exploitation of the currently available information and the exploration of new matches, we introduce the following UCB-based selection criterion.
Based on Lemma \ref{lemma:ucb}, the cumulative error induced by meta-learner is controlled by

\begin{displaymath}
\begin{aligned}
\sum_{t=1}^T & \underset{ r_t \mid \bx_t}{\bbe}  \bigg[ |f(\bx_{t}; \Theta_{t}) - r_{t}|  ~\big|~ u_t \bigg]  \\
& \leq  \sum_{t=1}^T    \underbrace{ \frac{\calo ( \| \nabla_{\Theta} f (\bx_{t}; \Theta_{t}) - \nabla_{\theta} f (\bx_{t}; \theta_{0}^{u_t})\|_2 )}{m^{1/4}} }_{\text{Meta-side info}}
\end{aligned}
\end{displaymath}

\begin{displaymath}
     \quad + \sum_{u \in N}  \mu_T^{u}   \Bigg  [ \underbrace{\calo \left( \sqrt{ \frac{S  + 1 } {2\mu^{u}_{T}}}  \right)+   \sqrt{\frac{  2 \log(  1 /\delta) }{\mu^{u}_T}}}_{\text{User-side info}} \Bigg],
\end{displaymath}
where $\nabla_{\Theta} f (\bx_{t}; \Theta_{t})$ incorporates the discriminative information of meta-learner acquired from the correlations within the relative cluster $\widehat{\caln}_{u_t}(\bx_{t})$ and $\calo(\frac{1}{\sqrt{\mu^{u}_T }})$ shows the shrinking confidence interval of user-learner to a specific user $u$. 
Then, we select an arm according to:
$
\bx_t = \arg_{ \bx_{t,i} \in \BX_t } \max   \mathbf{U}_{t,i}$ ( where $\mathbf{U}_{t,i}$ is calculated in Line 9).

In summary, Algorithm \ref{alg:main} depicts the workflow of \sysn. In each round $t$, given a target user and a pool of candidate arms, we compute the meta-learner and its bound for each relative cluster (Line 6-10). Then, we choose the arm according to the UCB-type strategy (Line 11). After receiving the reward, we update the user-learners. Note that the meta-learner has been updated in Line 8.

Then, we discuss the time complexity of Algorithm \ref{alg:main}.
Here, with $n$ being the number of users, \sysn will take $\calo(n)$ to find the cluster for the served user. Given the detected cluster $\widehat{\mathcal{N}}$, it takes $\calo( |\widehat{\mathcal{N}}|)$ to update the meta-learner by SGD. 
Suppose $\bbe[|\widehat{\mathcal{N}}|]=n/\hat{q}$ and $n/\hat{q} \ll n$. Therefore, the overall test time complexity of Algorithm \ref{alg:main} is $\calo(K (n + n/\hat{q}))$. 
To scale \sysn for deployment in large recommender systems, we can rely on the assistance of pre-processing tools: Pre-clustering of users and Pre-selection of items. On the one hand, we can perform pre-clustering of users based on the user features or other information. Then, let a pre-cluster (instead of a single user) hold a neural network, which will significantly reduce $n$. On the other hand, we can conduct the pre-selection of items based on item and user features, to reduce $K$ substantially. For instance, we only consider the restaurants that are near the serving user for the restaurant recommendation task.
Furthermore, we can also control the magnitude of $n/\hat{q}$ by tuning the hyperparameter $\nu$ based on the actual application scenario.
Consequently, \sysn can effectively serve as a core component of large-scale recommender systems.

\section{Regret Analysis} \label{sec:theo1}

In this section, we provide the performance guarantee of \sysn, which is built in the over-parameterized neural networks regime. 

As the standard setting in contextual bandits, all arms are normalized to the unit length.
Given an arm $\bx_{t,i} \in \bbr^d$ with $\|\bx_{t,i}\|_2 = 1$, $t \in [T], i \in [K]$,  without loss of generality, we define $f$ as a fully-connected network with depth $L \geq 2$ and width $m$:
\begin{equation} \label{eq:structure}
f(\bx_{t,i}; \theta \ \text{or} \ \Theta  ) = \bw_L \sigma ( \bw_{L-1}  \sigma (\bw_{L-2} \dots  \sigma(\bw_1 \bx_{t,i}) ))
\end{equation}
where $\sigma$ is the ReLU activation function,  $\bw_1 \in \bbr^{m \times d}$, $ \bw_l \in \bbr^{m \times m}$, for $2 \leq l \leq L-1$, $\bw^L \in \bbr^{1 \times m}$, and 
\[
\theta, \Theta  = [ \text{vec}(\bw_1)^\top,  \text{vec}(\bw_2)^\top, \dots, \text{vec}(\bw_L )^\top ]^\top \in \bbr^{p}.
\]
Note that our analysis results can also be readily generalized to other neural architectures such as CNNs and ResNet \cite{allen2019convergence, du2019gradient}.
Then, we employ the following initialization \citep{cao2019generalization} for $\theta$ and $\Theta$ : For $l \in [L-1]$, each entry of $\bw_l$ is drawn from the normal distribution $\caln(0, 2/m)$; Each entry of $\bw_L$ is drawn from the normal distribution $\caln(0, 1/m)$.
Here, given $R > 0$, we define the following function class:
\begin{equation}
    B(\theta_0, R) = \{ \theta \in \bbr^p:  \| \theta - \theta_0\|_2 \leq R/m^{1/4}\}.
\end{equation}
 The term $B(\theta_0, R)$ defines a function class ball centered at the random initialization point $\theta_0$ and with a radius of $R$. 
 This definition was originally introduced in the context of analyzing over-parameterized neural networks, and it can be found in the works of \citep{cao2019generalization} and  \citep{allen2019convergence}.
Recall that $q_{t,i}$ represents the number of clusters given $\bx_{t,i}$.
For the simplicity of analysis, we assume $\bbe[q_{t,i}] = q, t \in [T], i \in [K]$.
Let $\{(\bx_t, r_t)\}_{t=1}^{TK}$ represent all the data in $T$ rounds and define the squared loss $\call_t(\theta) = (f(\bx_t; \theta) - r_t)^2/2$. 
Then, we provide the instance-dependent regret upper bound for \sysn with the following theorem.  

\begin{restatable}{theorem}{theoremmain} \label{theorem:main}
Given the number of rounds $T$ and $\gamma$, for any $\delta \in (0, 1),  R > 0$, suppose $m \geq \widetilde{\Omega} ( \text{poly}(T, L, R) \cdot Kn\log (1/\delta))$, $ \eta_1 = \eta_2  = \frac{R^2}{\sqrt{m}}$, and $\bbe[|\caln_{u_t}(\bx_t)|] = \frac{n}{q}, t \in [T]$.
Then, with probability at least $1 - \delta$ over the initialization,  Algorithm \ref{alg:main} achieves the following regret upper bound:
\[
\mathbf{R}_T \leq \sqrt{  q T \cdot S^{\ast}_{TK}  + \calo(1) }    + \calo(\sqrt{   2 q T \log(\calo (1) /\delta)}).
\]
where $S^{\ast}_{TK} =  \underset{ \theta \in B(\theta_0, R)}{\inf} \sum_{t=1}^{TK} \call_t(\theta)$.
\end{restatable}

Theorem \ref{theorem:main} provides a regret bound for \sysn, which consists of two main terms. The first term is instance-dependent and relates to the squared error achieved by the function class $B(\theta_0, R)$ on the data. The second term is a standard large-deviation error term.

There are some noteworthy properties regarding Theorem \ref{theorem:main}. One important aspect is that it depends on the parameter $q$, which represents the expected number of clusters, rather than the number of users $n$. Specifically, $\wcalo(\sqrt{T})$ corresponds to the regret effort for learning a single bandit, and thus $\wcalo(\sqrt{nT})$ is an estimate of the regret effort for learning $n$ bandits. However, Theorem \ref{theorem:main} refines this naive bound to $\wcalo(\sqrt{qT})$, linking the regret effort to the actual underlying clusters among users.

Another advantage of Theorem \ref{theorem:main} is that it makes no assumptions about the contexts $\{\bx_t\}_{t=1}^{TK}$ used in the problem. This makes Theorem \ref{theorem:main} robust against adversarial attacks on the contexts and allows the observed contexts to contain repeated items.
In contrast, existing neural bandit algorithms like \cite{zhou2020neural, zhang2020neural, kassraie2022neural} rely on Assumption \ref{assum:ntk} for the contexts, and their regret upper bounds can be disrupted by straightforward adversarial attacks, e.g., creating two identical contexts with different rewards. 

The term $S^{\ast}_{TK}$ reflects the "regression difficulty" of fitting all the data using a given function class, while the radius $R$ controls the richness or complexity of that function class. It's important to note that the choice of $R$ is flexible, although it's not without constraints: specifically, the value of $m$ must be larger than a polynomial of $R$.
When $R$ is set to a larger value, it expands the function class $B(\theta_0, R)$, which means it can potentially fit a wider range of data. Consequently, this tends to make $S^{\ast}_{TK}$ smaller.
Recent advances in the convergence of neural networks, as demonstrated by \cite{allen2019convergence} and \cite{du2019gradient}, have shown that there is an optimal region around the initialization point in over-parameterized neural networks. This suggests that, with the proper choice of $R$, term $S^{\ast}_{TK}$ can be constrained to a small constant value.

Next, we show the common assumption made on existing neural bandits, and prove that  Theorem \ref{theorem:main} is no worse than their regret bounds under the same assumption.
The analysis is associated with the Neural Tangent Kernel (NTK) matrix as follows: 

\begin{definition} [NTK \cite{ntk2018neural, wang2021neural}] Let $\mathcal{N}$ denote the normal distribution.
Given the data instances $\{\bx_t\}_{t=1}^T$, for all $i, j \in [T]$,  define 
\[
\begin{aligned}
&\mathbf{H}_{i,j}^0 = \Sigma^{0}_{i,j} =  \langle \bx_i, \bx_j\rangle,   \ \ 
\mathbf{A}^{l}_{i,j} =
\begin{pmatrix}
\Sigma^{l}_{i,i} & \Sigma^{l}_{i,j} \\
\Sigma^{l}_{j,i} &  \Sigma^{l}_{j,j} 
\end{pmatrix} \\
&   \Sigma^{l}_{i,j} = 2 \mathbb{E}_{a, b \sim  \mathcal{N}(\mathbf{0}, \mathbf{A}_{i,j}^{l-1})}[ \sigma(a) \sigma(b)], \\ & \mathbf{H}_{i,j}^l = 2 \mathbf{H}_{i,j}^{l-1} \mathbb{E}_{a, b \sim \mathcal{N}(\mathbf{0}, \mathbf{A}_{i,j}^{l-1})}[ \sigma'(a) \sigma'(b)]  + \Sigma^{l}_{i,j}.
\end{aligned}
\]
Then, the NTK matrix is defined as $ \mathbf{H} =  (\mathbf{H}^L + \Sigma^{L})/2$.
\end{definition}

\begin{assumption} \label{assum:ntk}
There exists $\lambda_0 > 0$, such that $\bbh \succeq \lambda_0 \mathbf{I}$
\end{assumption}

The assumption \ref{assum:ntk} is generally held in the literature of neural bandits \cite{zhou2020neural,zhang2020neural,dai2022federated, jia2021learning, ban2021ee, ban2021multi, xu2020neural} to ensure the existence of a solution for NTK regression.
This assumption holds true when any two contexts in $\{\bx_t\}_{t=1}^{TK}$ are not linearly dependent or parallel. Then, the SOTA regret upper bound for a \textit{single} neural bandit ($n = 1$) \cite{zhou2020neural,zhang2020neural,dai2022federated, ban2021multi} is as follows:
\begin{equation}\label{eq:boundofneuralbandits}
 \wcalo(\sqrt{\hd  T} (S + \sqrt{\hd})).
\end{equation}
There are two complexity terms in the regret bounds \cite{zhou2020neural,ban2021ee}.
The first complexity term is $S = \sqrt{\bh^\top \mathbf{H}^{-1} \bh}$, where 
\[ \bh = [h_{u_1}(\bx_1), h_{u_1}(\bx_2), \dots, h_{u_T}(\bx_{TK}) ]^\top \in \bbr^{TK}.
\]
The purpose of the term $S$ is to provide an upper bound on the optimal parameters in the context of NTK regression. However, it's important to note that the value of $S$ becomes unbounded (i.e., $\infty$) when the matrix $\mathbf{H}$ becomes singular. This singularity can be induced by an adversary who creates two identical or parallel contexts, causing problems in their analysis.

The second complexity term is the effective dimension $\tilde{d}$, defined as $ \hd = \frac{\log \det(\mathbf{I} + \mathbf{H} )}{\log (1 + TK)}$, which describes the actual underlying dimension in the RKHS space spanned by NTK. 
The following lemma is to show an upper bound of $S^{\ast}_{TK}$ under the same assumption.

\begin{lemma} \label{lemma:upperboundofStk}
Suppose Assumption \ref{assum:ntk} and conditions in Theorem \ref{theorem:main} holds where $m \geq \widetilde{\Omega} ( \text{poly}(T, L) \cdot Kn \lambda_0^{-1} \log (1/\delta))$. With probability at least $1 - \delta$ over the initialization, there exists $\btheta'  \in  B(\theta_0, \widetilde{\Omega}(T^{3/2}))$, such that
\[
\begin{aligned}
\bbe[S^{\ast}_{TK}] \leq   \bbe [ \sum_{t=1}^{TK}  \call_t(\theta')   ] \leq  \wcalo \left(\sqrt{ \widetilde{d}} + S  \right)^2 \cdot \widetilde{d}.
\end{aligned}
\]
\end{lemma}
Lemma \ref{lemma:upperboundofStk} provides an upper bound for $S^{\ast}_{TK}$ by setting $R =  \widetilde{\Omega}(T^{3/2})$. Subsequently, by applying the Hoeffding-Azuma inequality over $S^{\ast}_{TK}$ and replacing $S^{\ast}_{TK}$ with this upper bound, Theorem \ref{theorem:main} can be reformulated as $\wcalo(\sqrt{\hd T} (S + \sqrt{\hd}))$ for a single neural bandit or $ \wcalo(\sqrt{q\hd T} (S + \sqrt{\hd}))$ for $n$ users (CNB problem). This transformation implies that Theorem \ref{theorem:main} is at least as good as the SOTA upper bounds represented by Eq. \eqref{eq:boundofneuralbandits}.

\section{Experiments} \label{sec:exp}

In this section, we evaluate \sysn's empirical performance on both online recommendation and classification scenarios. Our source code are anonymously available at \textit{\url{https://anonymous.4open.science/r/Mn-C35C/}}.

\textbf{Recommendation datasets}. We use four public datasets,
Amazon \citep{ni2019justifying}, Facebook \cite{leskovec2012learning},
Movielens \citep{harper2015movielens},  and Yelp  \footnote{\url{https://www.yelp.com/dataset}}, to evaluate \sysn's ability in discovering and exploiting user clusters to improve the recommendation performance. 
Amazon is an E-commerce recommendation dataset consisting of $883636$ review ratings. 
Facebook is a social recommendation dataset with $88234$ links. 
MovieLens is a movie recommendation dataset consisting of $25$ million reviews between $1.6 \times 10^5$ users and $6 \times 10^4$ movies. 
Yelp is a shop recommendation dataset released in the Yelp dataset challenge, composed of 4.7 million review entries made by $1.18$ million users towards $1.57 \times 10^5$ merchants.
For these four datasets, we extract ratings in the reviews and build the rating matrix by selecting the top $10000$ users and top $10000$ items (friends, movies, shops) with the most rating records.
Then, we use the singular-value decomposition (SVD) to extract a normalized $10$-dimensional feature vector for each user and item.
The goal of this problem is to select the item with good ratings.
Given an item and a specific user, we generate the reward by using the user's rating stars for this item. If the user's rating is more than 4 stars (5 stars total), its reward is $1$; Otherwise, its reward is $0$. Here, we use pre-clustering (K-means) to form the user pool with 50 users (pre-clusters).
Then, in each round, a user $u_t$ is randomly drawn from the user pool.
For the arm pool,  we randomly choose one restaurant (movie) rated by $u_t$ with reward $1$ and randomly pick the other $9$ restaurants (movies) rated by $u_t$ with $0$ reward. With each restaurant or movie corresponding to an arm, the goal for the learner is to pick the arm with the highest reward.

\textbf{Classification datasets}.
In our online classification with bandit feedback experiments, we utilized a range of well-known classification datasets, including Mnist \citep{lecun1998gradient}, Notmnist \citep{bulatov2011notmnist}, Cifar10 \citep{krizhevsky2009learning}, Emnist (Letter) \citep{cohen2017emnist}, Fashion \citep{xiao2017online}, as well as the Shuttle, Mushroom, and MagicTelescope (MT) datasets \citep{asuncion2007uci}.
Here, we provide some preliminaries for this setup.
In the round $t \in [T]$, given an instance $\bx_t \in \bbr^d$ drawn from some distribution, we aim to classify $\bx_t$ among $K$ classes.  $\bx_t$ is first transformed into $K$ long vectors: $\bx_{t,1} = (\bx^\top, 0, \dots, 0)^\top, \bx_{t,2} = (0, \bx^\top, \dots, 0)^\top, \dots, \bx_{t,K} = (0, 0, \dots, \bx^\top)^\top \in \bbr^{dK}$, matching $K$ classes respectively. 
The index of the arm that the learner selects is the class predicted by the learner. Then, the reward is defined as $1$ if $\bx_t$ belongs to this class; otherwise, the reward is $0$. 
In other words, each arm represents a specific class. For example, $\bx_{t,1}$ is only presented to Class 1; $\bx_{t,2}$ is only presented to Class 2. 
This problem has been studied in almost all the neural bandit works \citep{zhou2020neural, zhang2020neural, kassraie2022neural, ban2021ee}. Compared to these works, we aim to learn the correlations among classes to improve performance. Thus, we formulate one class as a user (bandit) (i.e., a user in the recommendation scenario) and all the samples belonging to this class are deemed as the data of this user.
This set of experiments aims to evaluate \sysn's ability to learn various non-linear reward functions, as well as the ability of discovering and exploiting the correlations among classes.
Additionally, we extended the evaluation by combining the Mnist and Notmnist datasets to simulate a more challenging application scenario, given that both datasets involve 10-class classification problems.


\textbf{Baselines}. We compare \sysn with SOTA baselines as follows:
(1)  CLUB \citep{2014onlinecluster} clusters users based on the connected components in the user graph and refines the groups incrementally;
(2) COFIBA \citep{2016collaborative} clusters on both the user and arm sides based on the evolving graph, and chooses arms using a UCB-based exploration strategy;
(3) SCLUB \citep{2019improved} improves the algorithm CLUB by allowing groups to merge and split, to enhance the group representation;
(4) LOCB \citep{ban2021local} uses the seed-based clustering and allows groups to be overlapped. Then, it chooses the best group candidates for arm selection;
(5) NeuUCB-ONE \citep{zhou2020neural} uses one neural network to formulate all users, and selects arms via a UCB-based recommendation;
(6) NeuUCB-IND \citep{zhou2020neural} uses one neural network to formulate one user separately (totally $n$ networks) and applies the same strategy to choose arms.
(7) NeuA+U: we concatenate the arm features and user features together and treat them as the input for the neural network. Note that the user features are only available on Movielens and Yelp datasets. Thus, we only report the results on these two datasets for NeuA+U. 
(8) NeuralLinear: following the existing work \citep{nabati2021online, zahavy2019deep}. A shared neural network is built for all users to get an embedding for each arm. which is fed into the linear bandit with the clustering procedure.
Since LinUCB \citep{2010contextual} and KernalUCB \citep{valko2013finite} are outperformed by the above baselines, we will not include them for comparison.

\begin{figure}[t]
\centering
\includegraphics[width=.49\columnwidth]{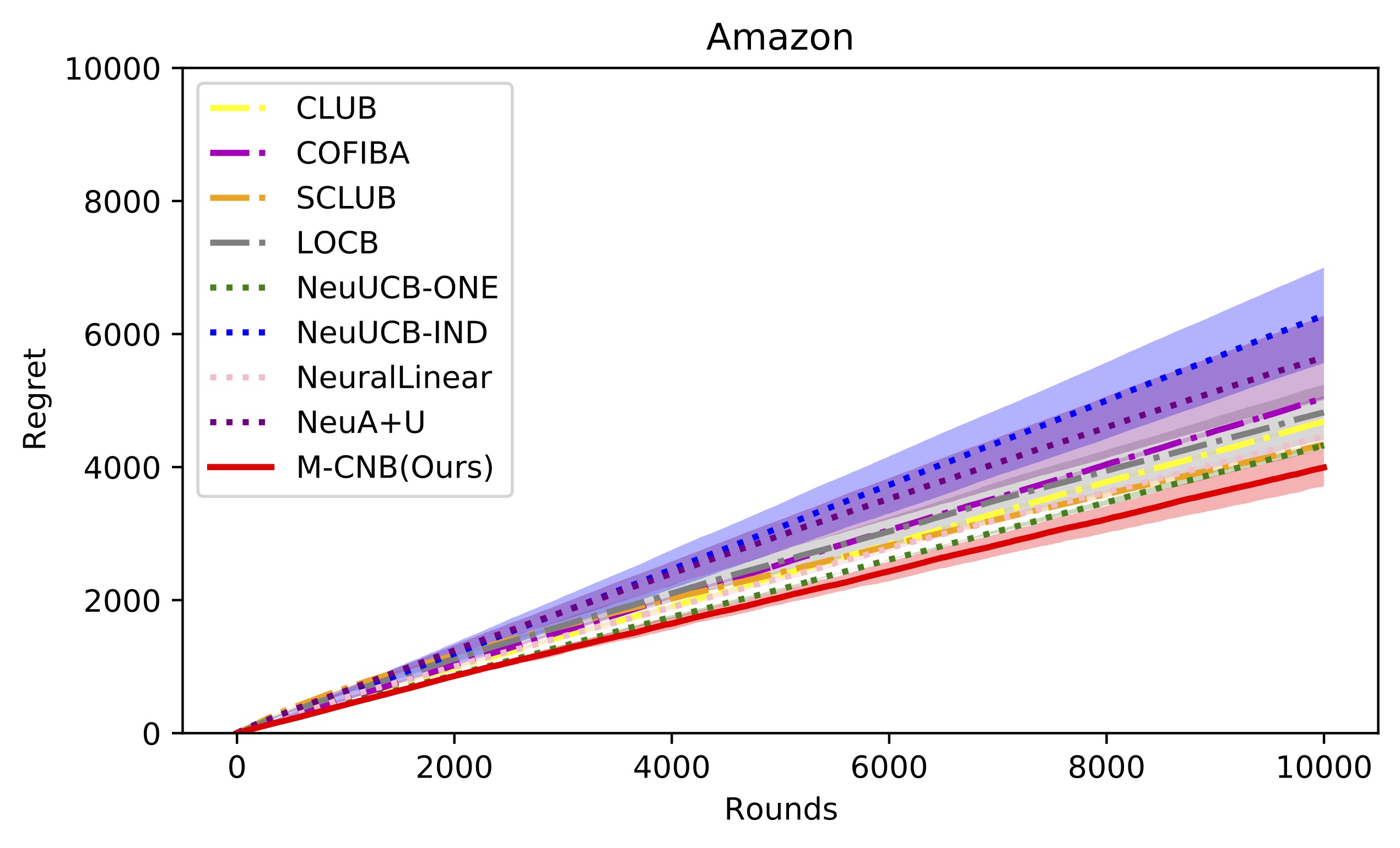}
    \includegraphics[width=.49\columnwidth]{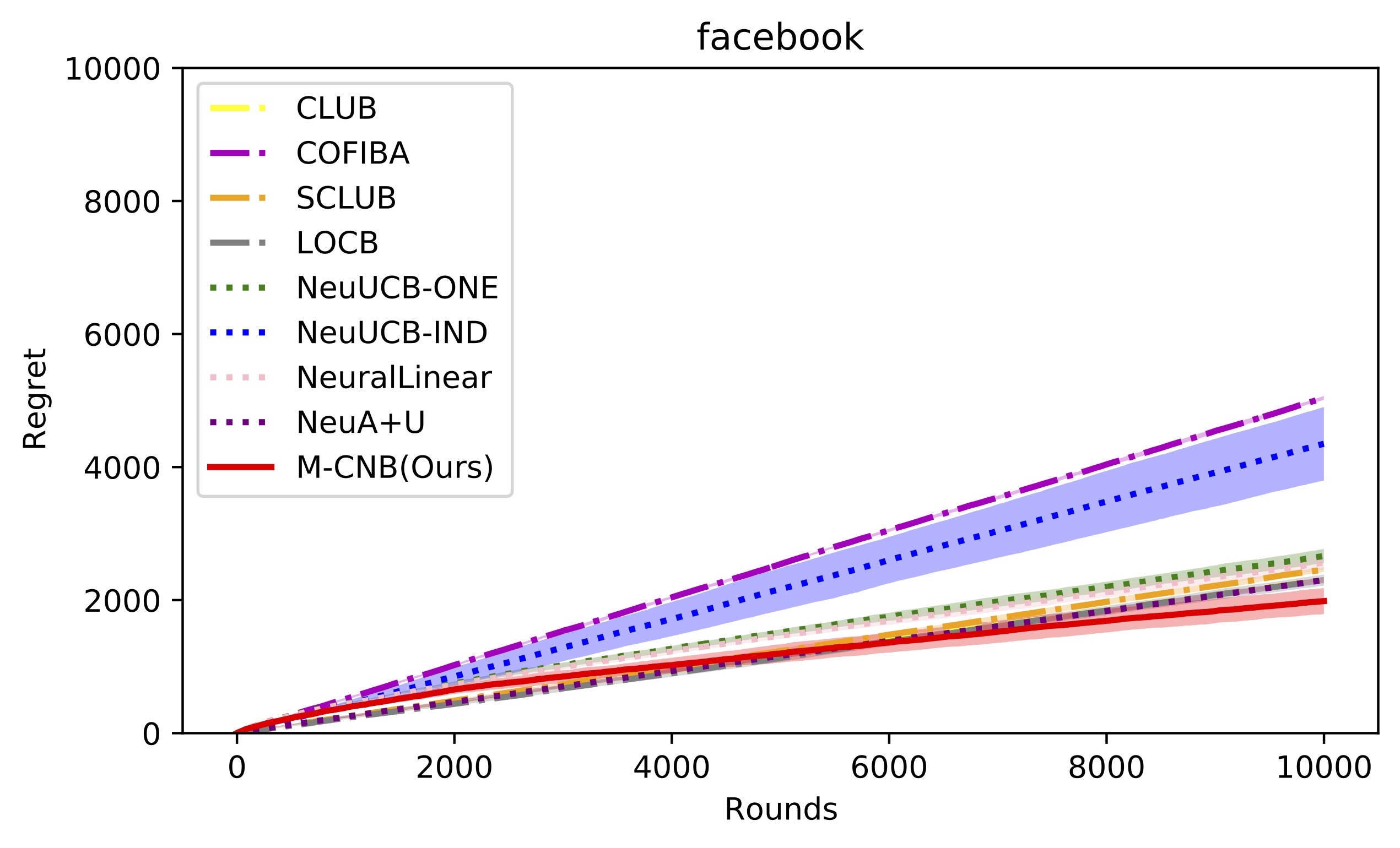}
    \includegraphics[width=.49\columnwidth]{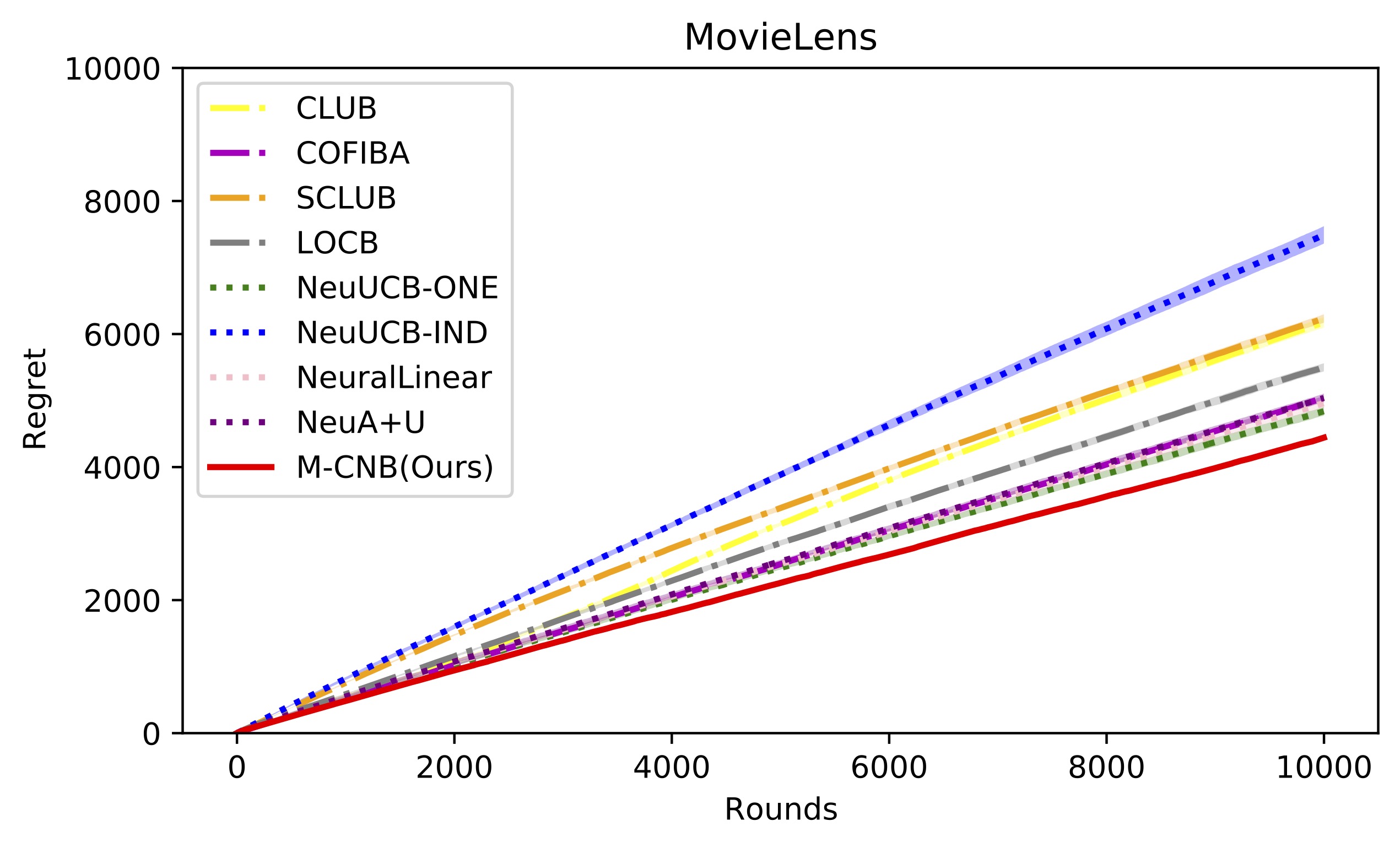}
    \includegraphics[width=.49\columnwidth]{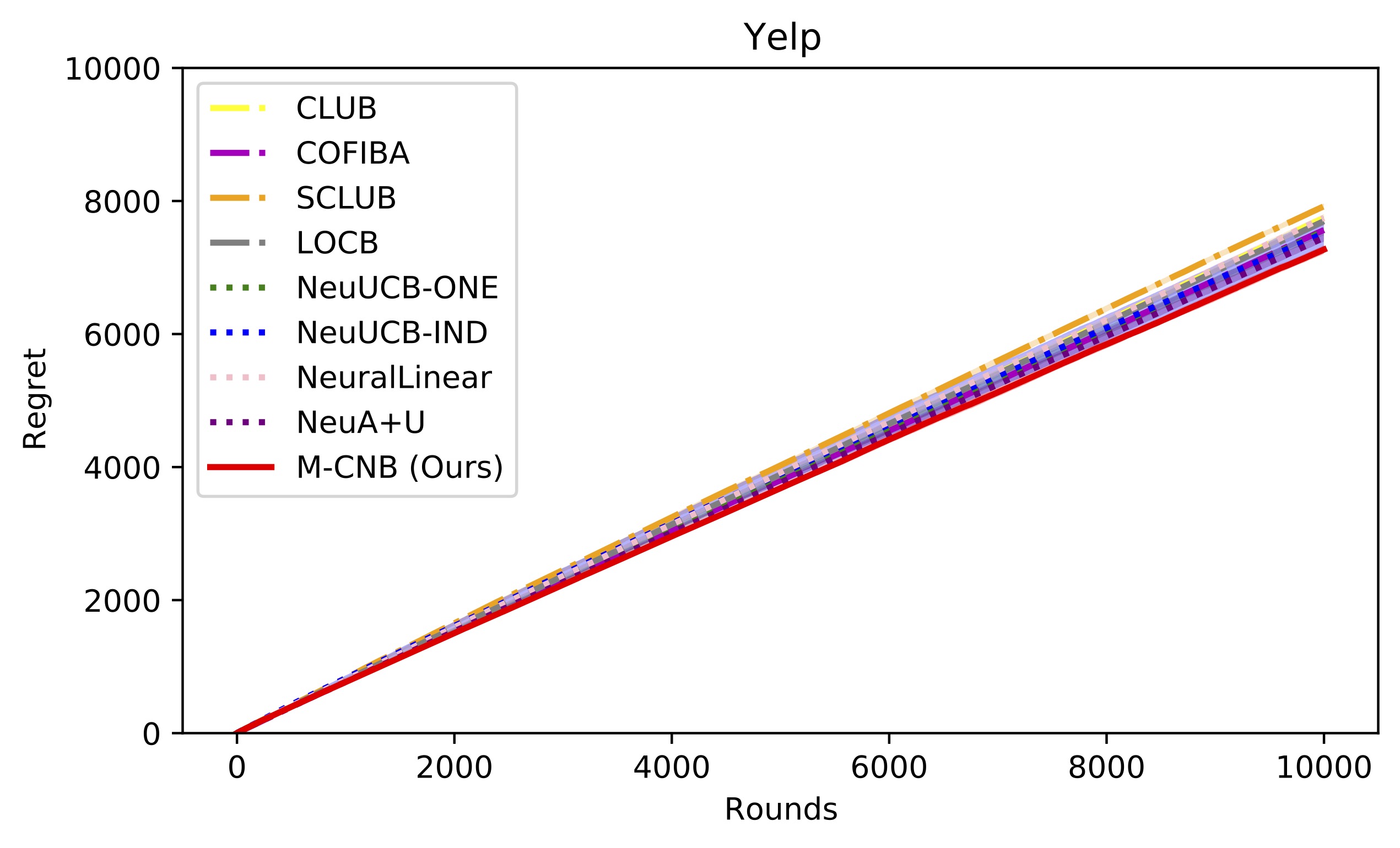}
   \caption{ Regret comparison on recommendation datasets. }
       \label{fig:recom}
\end{figure}


\textbf{Configurations}. 
We run all experiments on a server with the NVIDIA Tesla V100 SXM2 GPU.
For all the baselines, they all have two parameters: $\lambda$ that is to tune the regularization at initialization and $\alpha$ which is to adjust the UCB value. To find their best performance, we conduct the grid search for $\lambda$ and $\alpha$ over $(0.01, 0.1, 1)$ and $(0.0001, 0.001, 0.01, 0.1)$ respectively. 
For LOCB, the number of random seeds is set as $20$ following their default setting. For $\sysn$, we set $\nu$ as $5$ and $\gamma$ as $0.4$ to tune the cluster, and $S$ is set to $1$. 
To ensure fair comparison, for all neural methods, we use the same simple neural network with $2$ fully-connected layers, and the width $m$ is set as $100$.
To save the running time, we train the neural networks every 10 rounds in the first 1000 rounds and train the neural networks every 100 rounds afterwards.
In our implementation, we use Adam \cite{kingma2014adam} for SGD.
In the end,  we choose the best results for the comparison and report the mean and standard deviation (shadows in figures) of $10$ runs for all methods.

\begin{figure}[t]
\centering
    \includegraphics[width=.49\columnwidth]{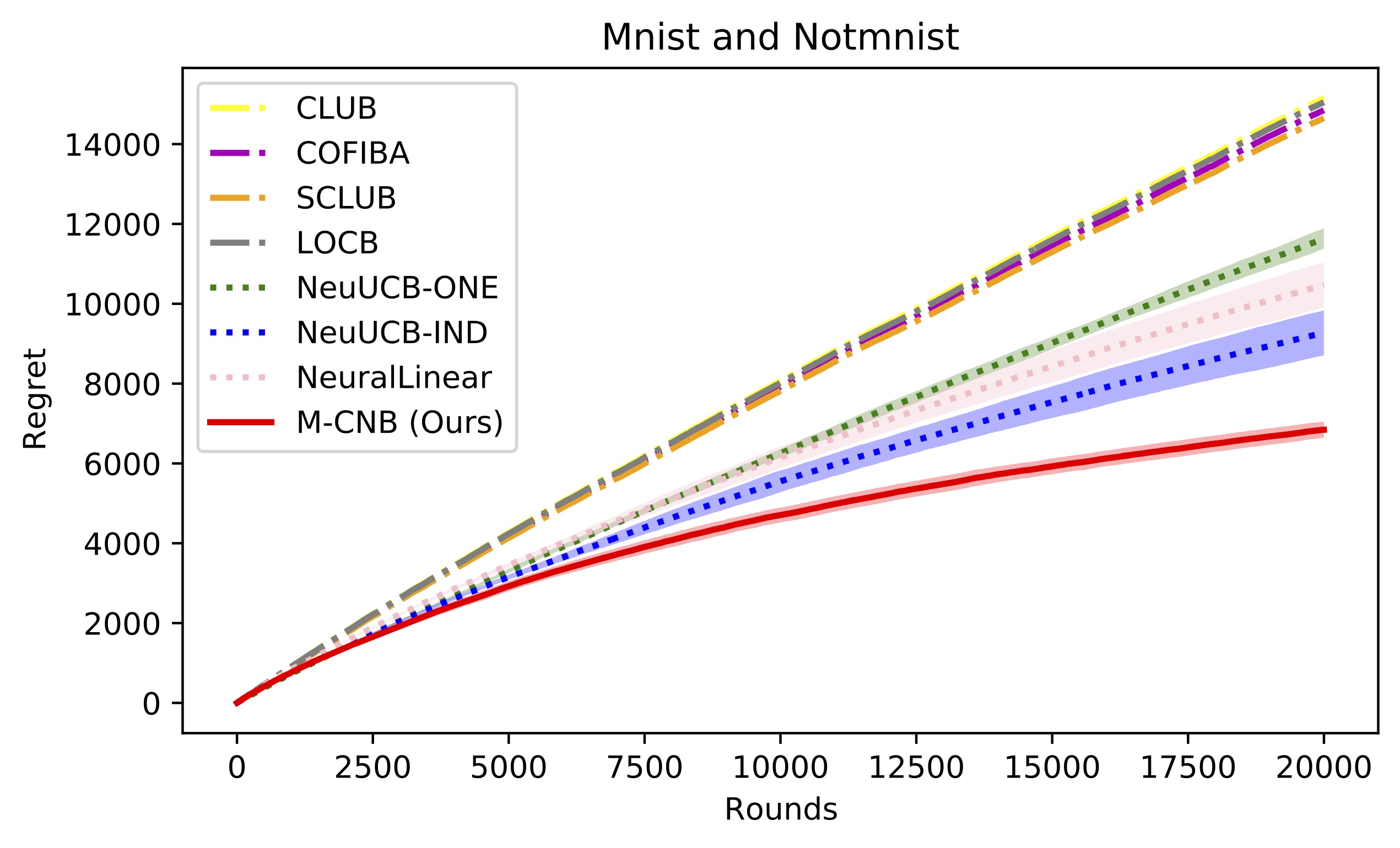}
    \includegraphics[width=.49\columnwidth]{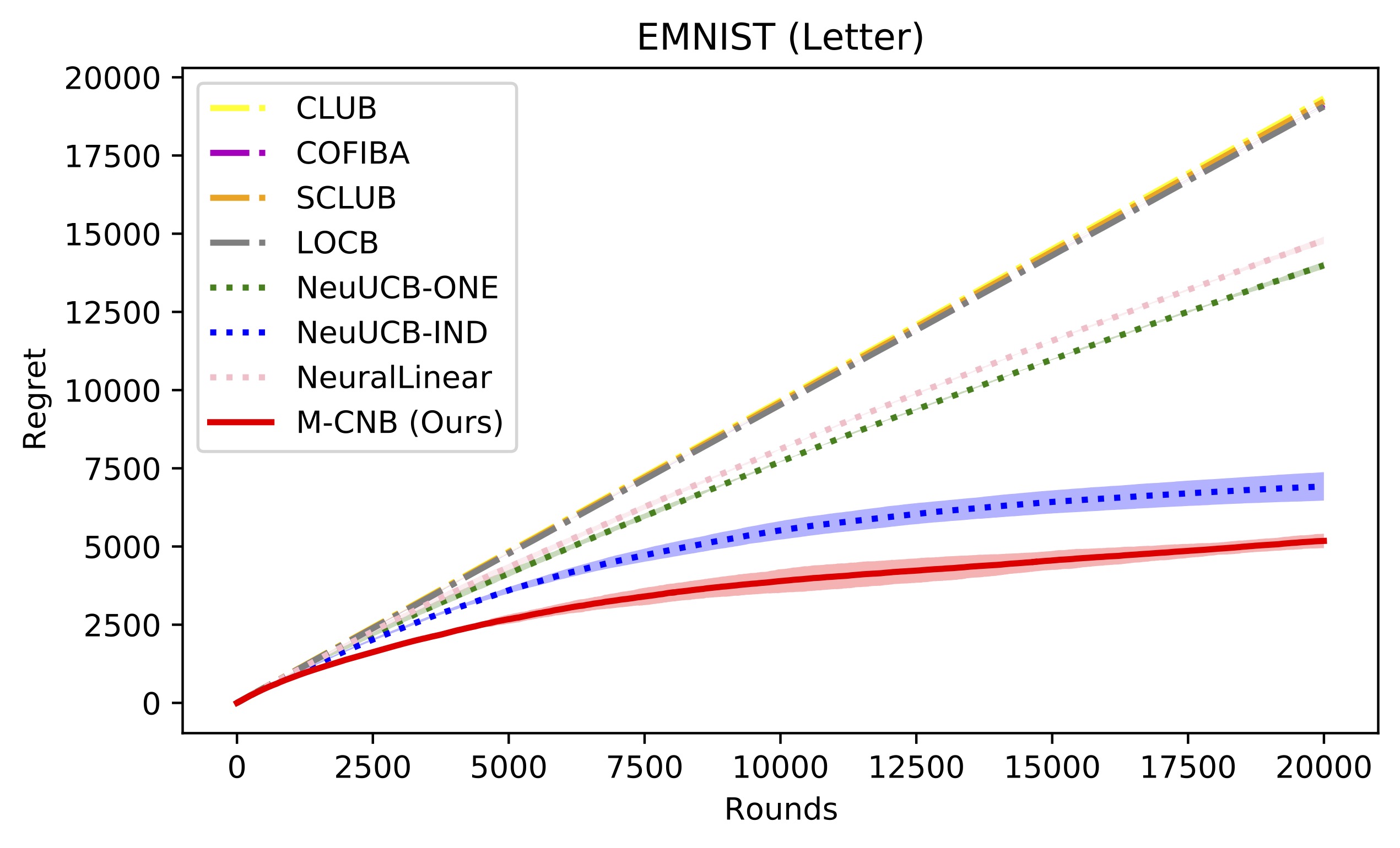}
    \includegraphics[width=.49\columnwidth]{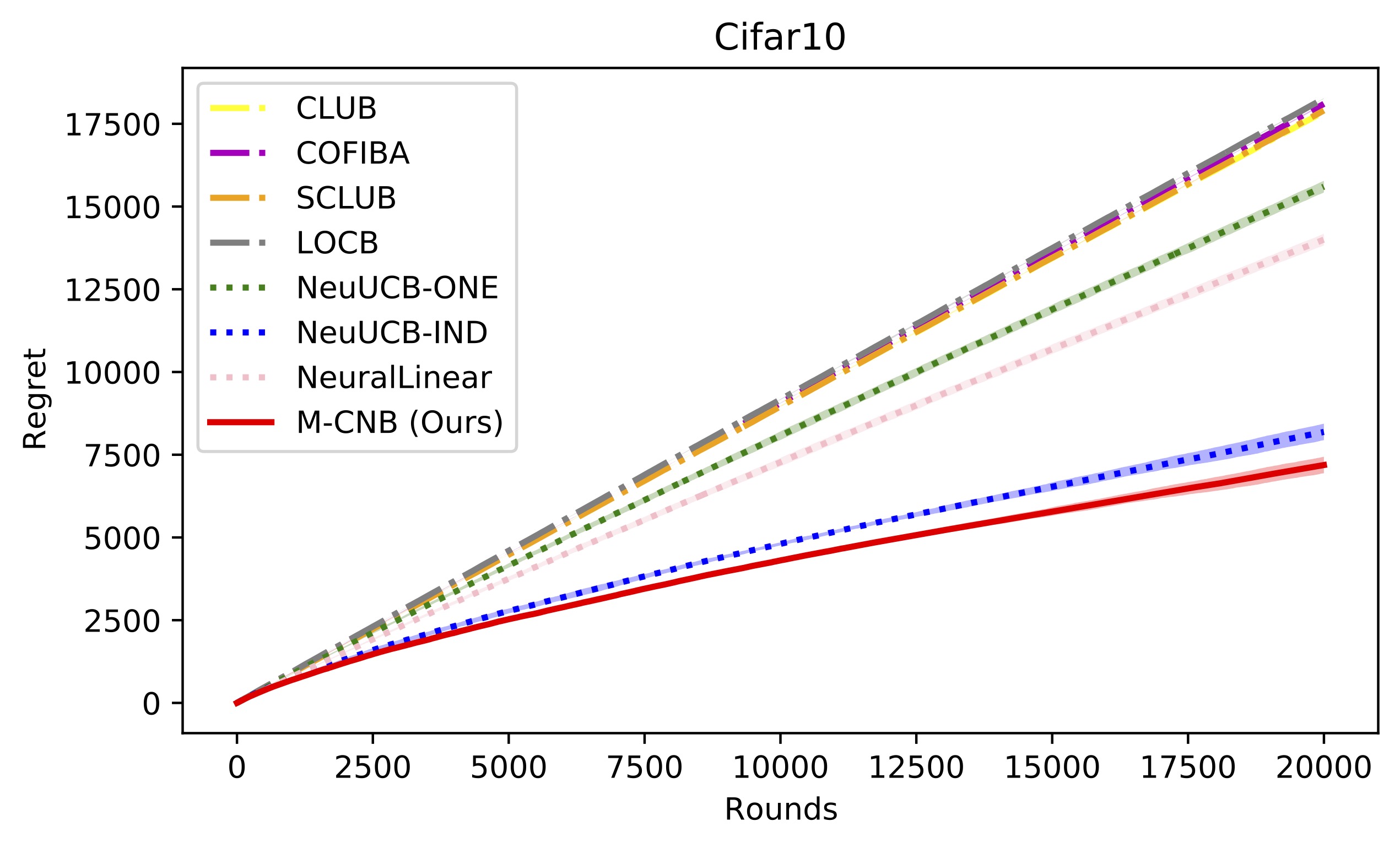}
    \includegraphics[width=.49\columnwidth]{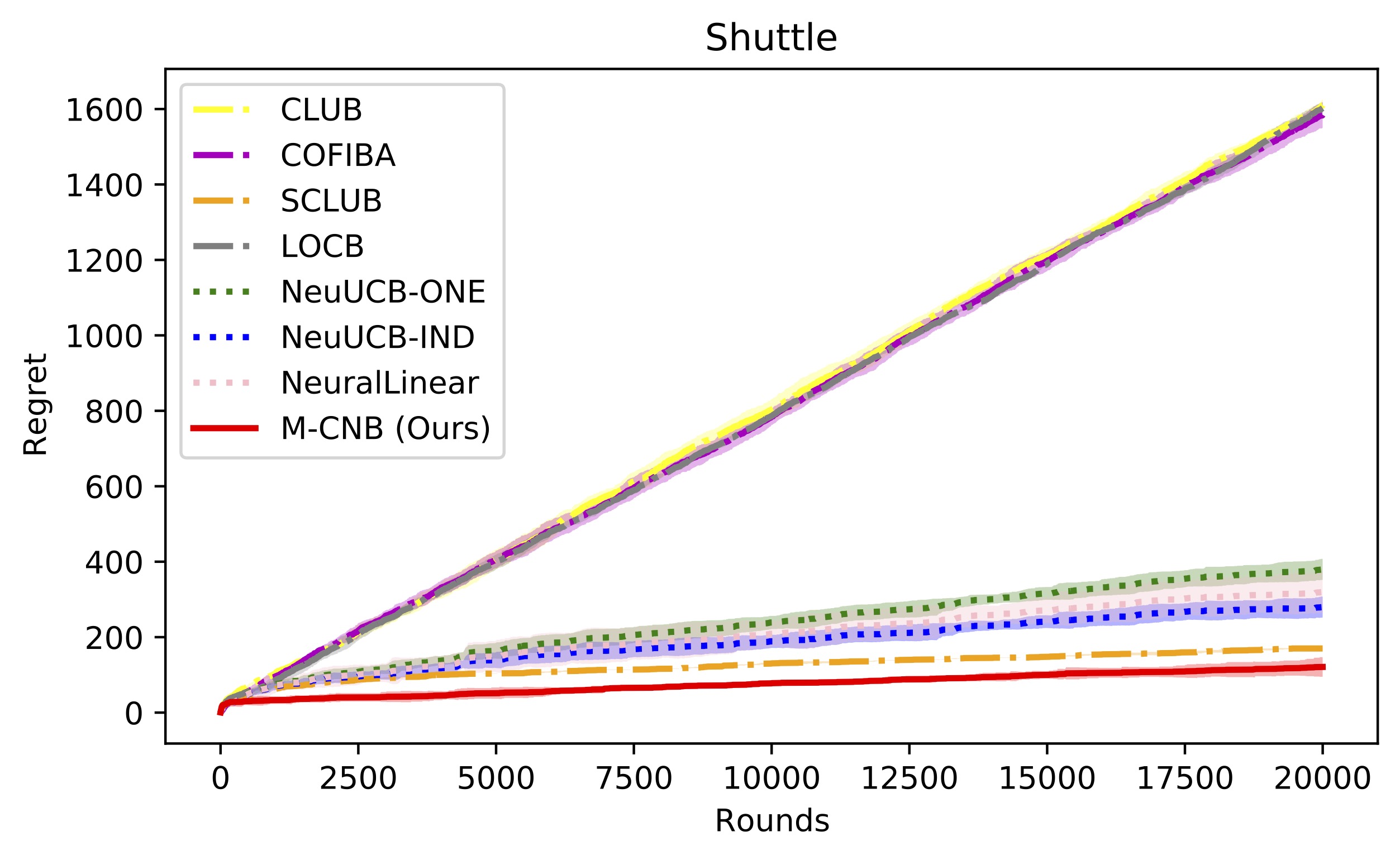}
\vspace{-0.1in}
\caption{ Regret comparison on  Mnist and Notmnist,  Cifar10,  EMNIST(Letter), and Shuttle. }
       \label{fig:mld}
\end{figure}

\begin{figure}[t]
\centering
    \includegraphics[width=.49\columnwidth]{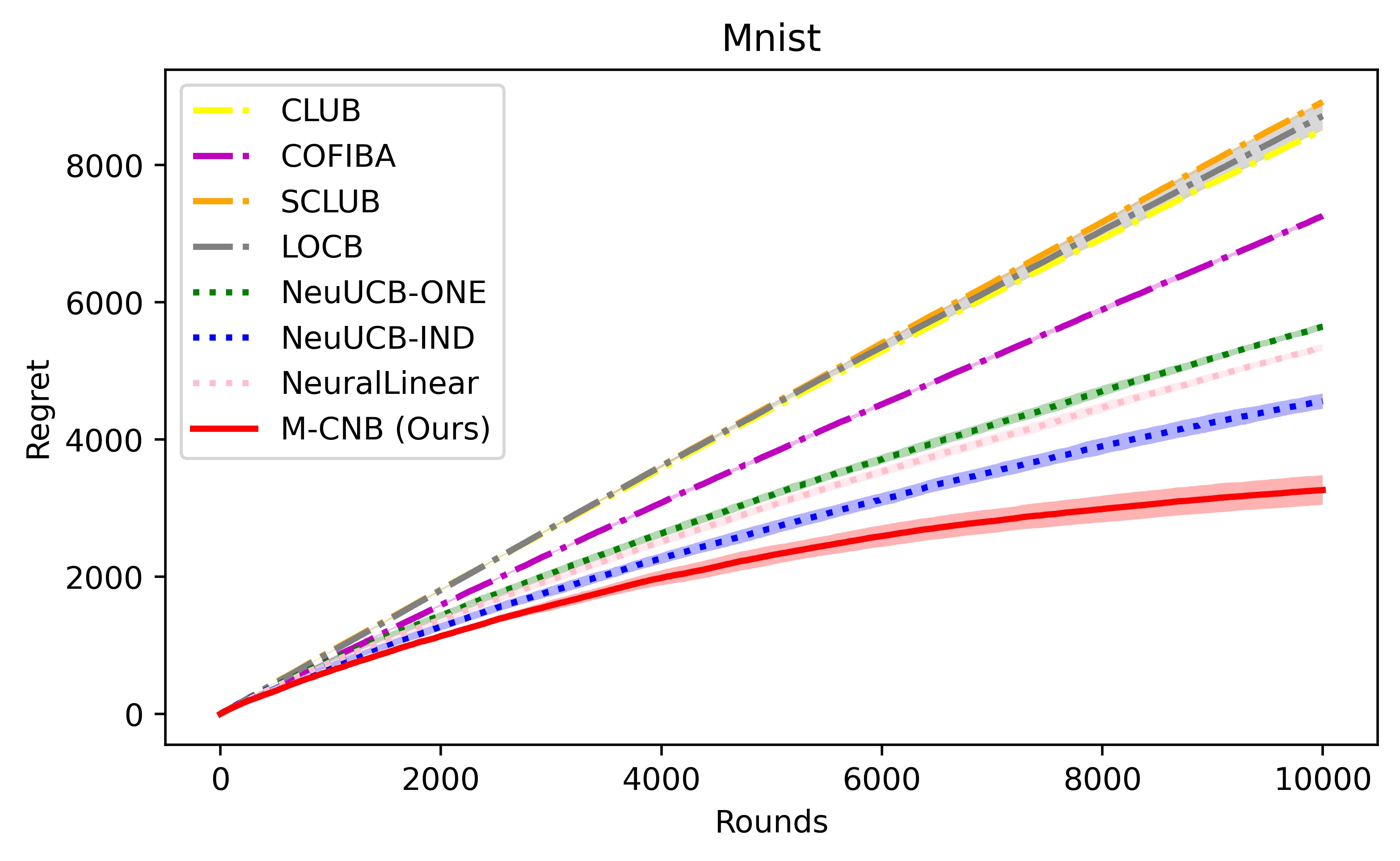} 
    \includegraphics[width=.49\columnwidth]{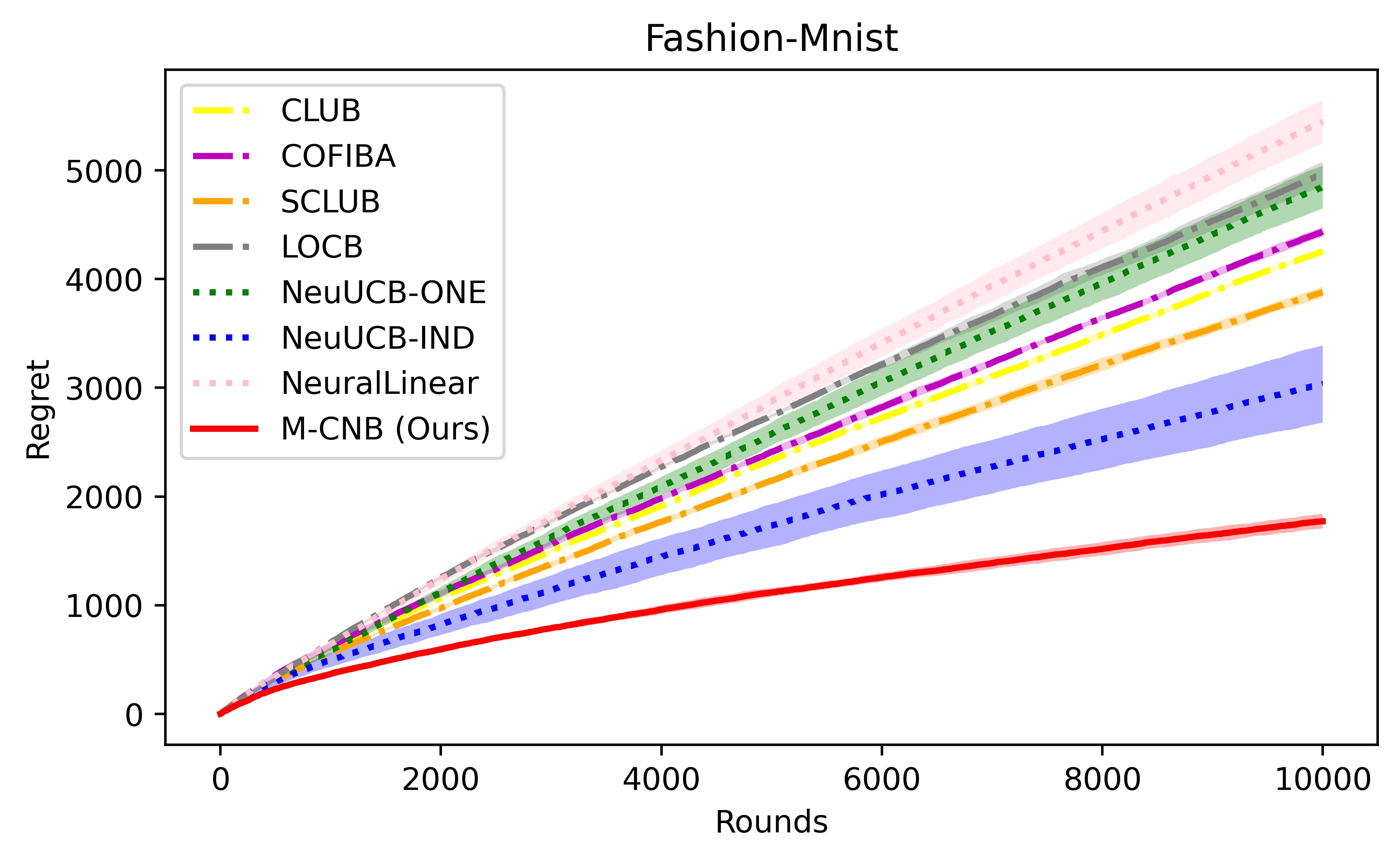}
    \includegraphics[width=.49\columnwidth]{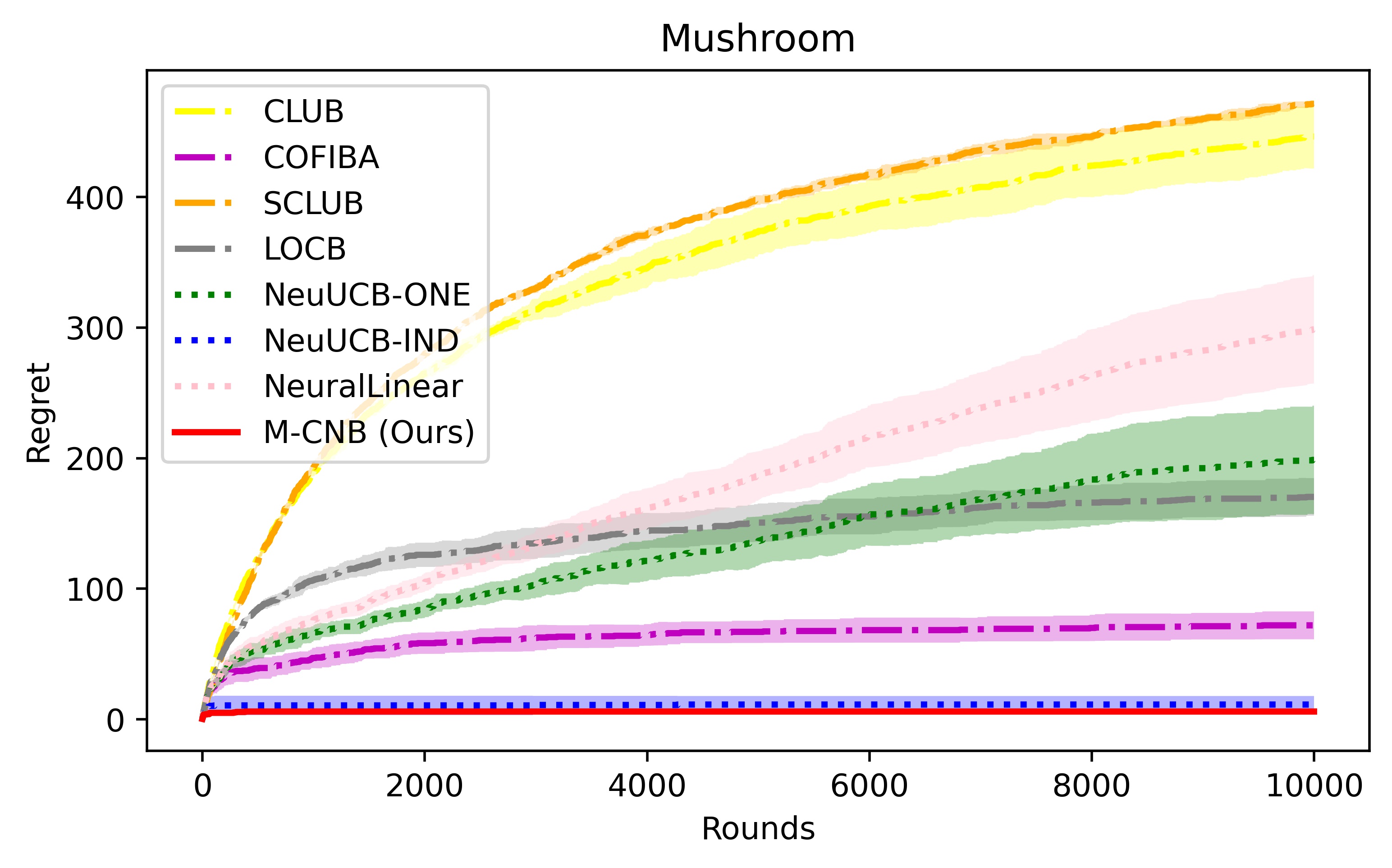}
    \includegraphics[width=.49\columnwidth]{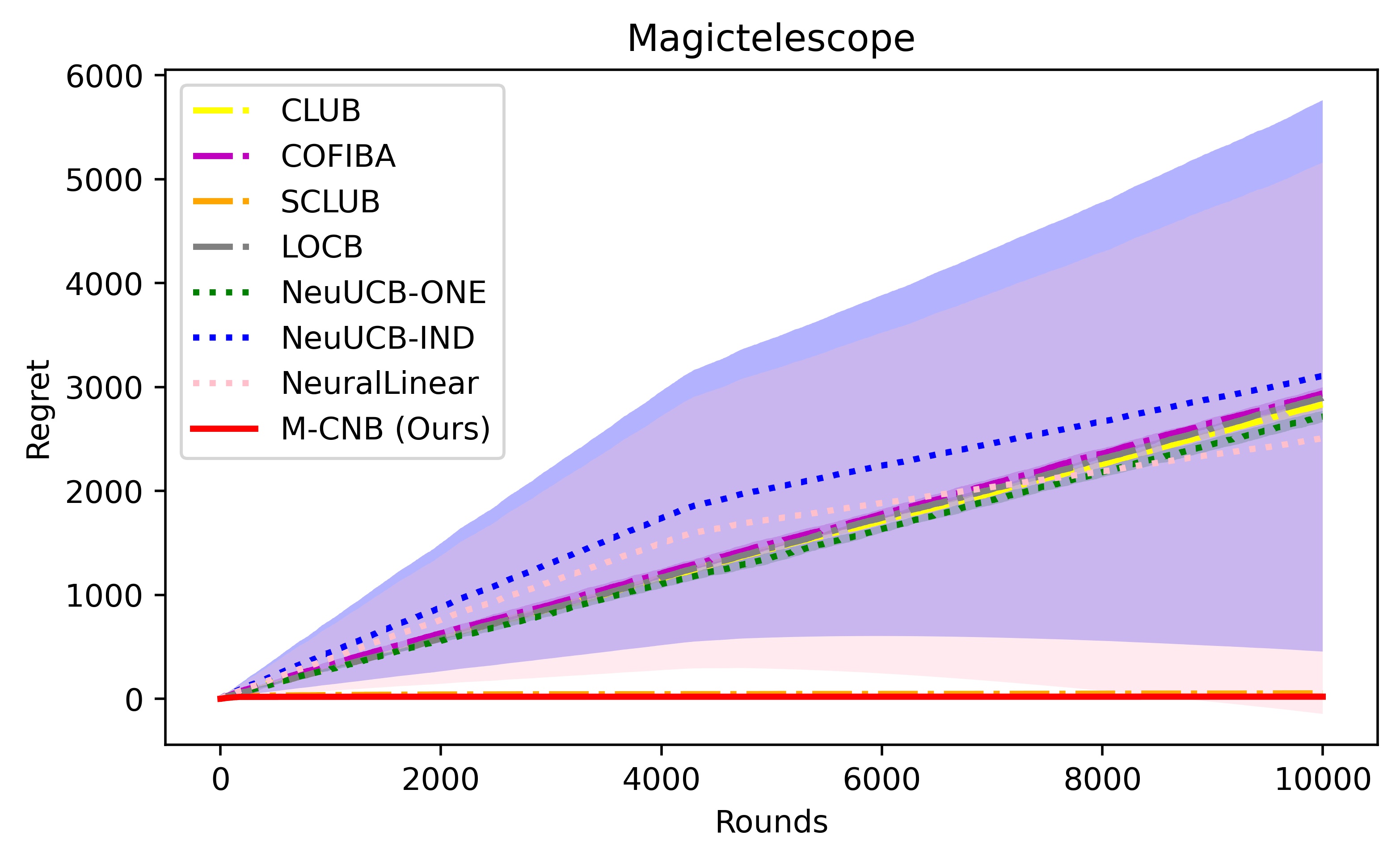}
\vspace{-0.1in}
\caption{ Regret comparison on  Mnist,  Fashion-Mnist, Mushroom, and MagicTelescope.
    }
       \label{fig:add}
\end{figure}

\textbf{Results}. 
Figure \ref{fig:recom}-\ref{fig:add} reports the average regrets of all the methods on the recommendation and classification datasets.
Figure \ref{fig:recom} displays the regret curves for all the methods evaluated on the MovieLens and Yelp datasets. In these experiments, \sysn consistently outperforms all the baseline methods, showcasing its effectiveness. Specifically, \sysn improves performance by 5.8\% on Amazon, 7.7 \% on Facebook, 8.1 \% on MovieLens, and 2.0 \% on Yelp, compared to the best-performing baseline.
These superior results can be attributed to two specific advantages that \sysn offers over the two types of baseline methods. In contrast to conventional linear clustering of bandits (CLUB, COFIBA, SCLUB, LOCB), \sysn has the capability to learn non-linear reward functions. This flexibility allows \sysn to excel in scenarios where user preferences exhibit non-linearity in terms of arm contexts. In comparison to neural bandits (NeuUCB-ONE, NeuUCB-IND, NeuA+U, NeuralLinear), \sysn takes advantage of user clustering and leverages the correlations within these clusters, as captured by the meta-learner. This exploitation of inter-user correlations enables \sysn to enhance recommendation performance.
By combining these advantages, \sysn achieves substantial improvements over the MovieLens and Yelp datasets, demonstrating its prowess in addressing collaborative neural bandit problems and enhancing recommendation systems.
Note \sysn's regret rate decreases on these four datasets, even though the "linear-like" behavior in Figure \ref{fig:recom}. 

Figures \ref{fig:mld} and \ref{fig:add} show the regret comparison on ML datasets, where \sysn outperforms all the baselines. Here, each class can be thought of as a user in these datasets.
The ML datasets exhibit non-linear reward functions concerning the arms, making them challenging for conventional  clustering of linear bandits (CLUB, COFIBA, SCLUB, LOCB). These methods may struggle to capture the non-linearity of the reward functions, resulting in sub-optimal performance.
Among the neural baselines, NeuUCB-ONE benefits from the representation power of neural networks. However, it treats all users (classes) as a single cluster, overlooking the variations and correlations among them. On the other hand, NeuUCB-IND deals with users individually, neglecting the potential benefits of leveraging collaborative knowledge among users.
NeuralLinear uses one shared embedding (neural network) for all users, which may not be the optimal solution given the user heterogeneity.
\sysn's advantage lies in its ability to exploit shared knowledge within clusters of classes that exhibit strong correlations. It leverages this common knowledge to improve its performances across different tasks, as it can efficiently adapt its meta-learner based on past clusters.

\begin{figure}[ht]
\centering
    \includegraphics[width=.49\columnwidth]{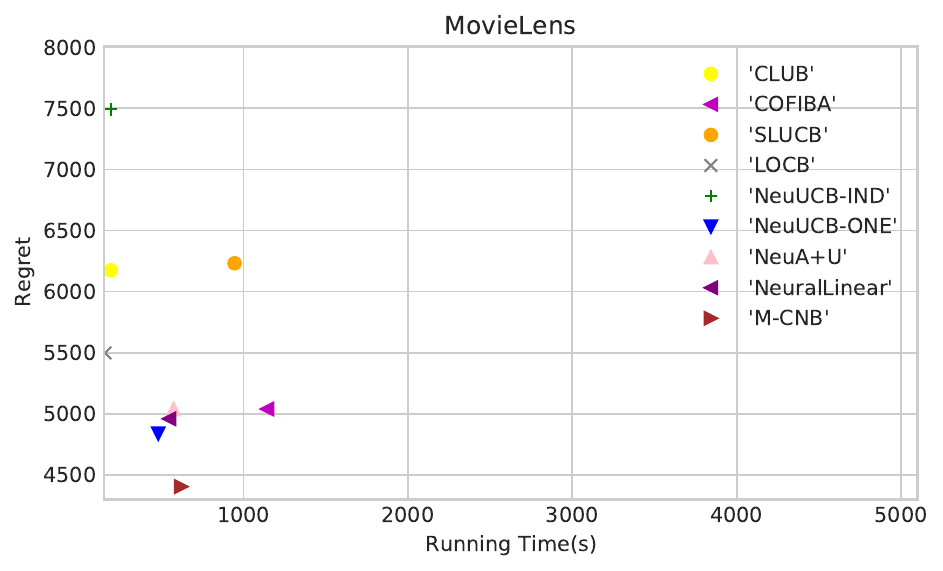}
    \includegraphics[width=.49\columnwidth]{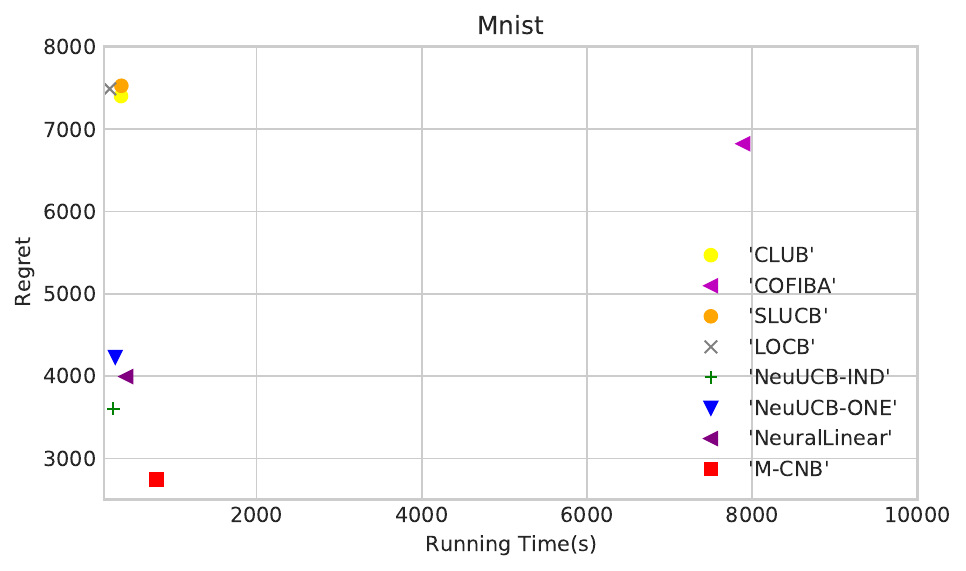}
   \caption{ Running time vs. Performance for all methods.  }
       \label{fig:runningtimeperformance}
\vspace{-0.1in}
\end{figure}


\textbf{Running time analysis}. 
Figure \ref{fig:runningtimeperformance} demonstrates the trade-off between running time and cumulative regret on both the Movielens and Mnist datasets, where the unit of the x-axis is seconds. As \sysn is under the framework of neural bandits, we use NeuUCB-ONE as the baseline (1.0).  The results indicate that \sysn takes comparable computation costs ($1.6 \times$ on Movielens and $2.9 \times$ on Mnist) to NeuUCB-ONE while substantially improving performance. 
This suggests that \sysn can be deployed to significantly enhance performance when the user correlation is a crucial factor (e.g., recommendation tasks), with only a moderate increase in computational overhead.

Now, let us delve into the analysis of the running time for M-CNB. 
Specifically, we can break down the computational cost of \sysn into three main components:
(1) \emph{Clustering}: to form the user cluster (Line 7 in Algorithm \ref{alg:main}); (2) \emph{Meta adaptation}: to train a meta-model (Algorithm \ref{alg:meta}); (3) \emph{User-learner training}: to train the user-learners (Lines 14-18 in Algorithm \ref{alg:main}).

\begin{table}[h]
	\caption{  Breakdown time cost for \sysn in a round (seconds) with different number of users on MovieLens. }
	\vspace{-1em}
	\centering
 \begin{adjustbox}{width= 0.8\columnwidth,center}

	\begin{tabular}{c|c|c|c|c}
		\toprule 
            & n =500 & n = 5000 & n = 10000 & n = 20000      \\
             \midrule  
 Clustering &  0.006   &  0.057 & 0.113    &  0.228      \\         
 \midrule  
 Meta adaptation & 0.003 & 0.002 & 0.003&  0.003  \\                  \midrule  

User-learner training&   0.067  & 0.068 &  0.096 &   0.078 \\
\bottomrule
	\end{tabular} \label{tab:timeparts}
 \end{adjustbox}
\end{table}

Table \ref{tab:timeparts} provides the breakdown of the time cost for the three main components of \sysn.
Clustering: This part's time cost grows linearly with the number of users $n$ because it has a time complexity of $O(n)$ for clustering. As discussed previously, leveraging pre-clustering techniques can significantly reduce this cost. It is also important to note that all clustering methods inherently have this time cost, and it is challenging to further reduce it. 
Meta adaptation: Due to the benefits of meta-learning, this part requires only a few steps of gradient descent to train a model with good performances. Consequently, the time cost for meta-adaptation is relatively trivial.
User-learner training: While this part may require more SGD steps to converge, it is important to recognize that it is primarily used for clustering purposes. Therefore, the frequency of training user-learners can be reduced to decrease the cost. 
In summary, \sysn aims to achieve the clustering of neural bandits and can manage to strike a good balance between the computational cost and the model performance.


\begin{figure}[ht] 
    \centering
    \includegraphics[width=0.49\columnwidth]{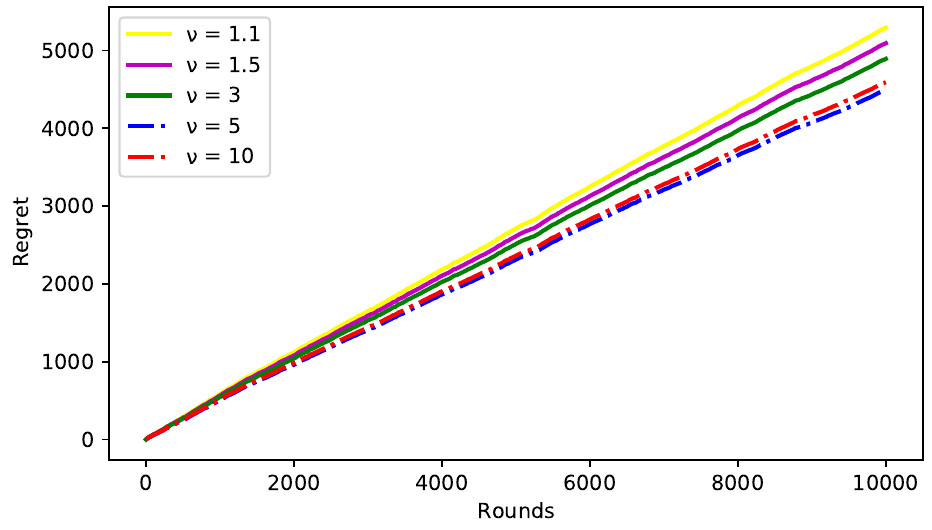}
    \includegraphics[width=0.49\columnwidth]{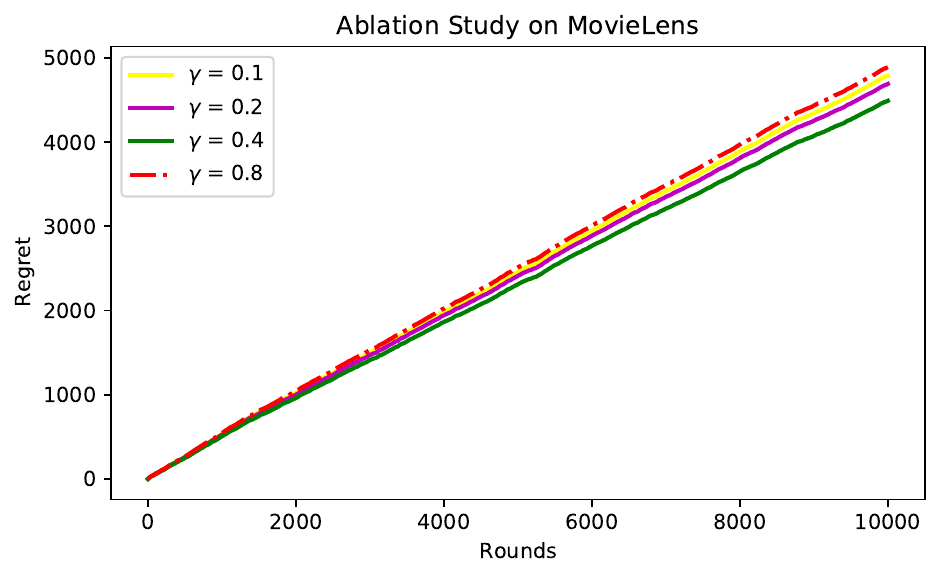}\hfill
    \centering
     \vspace{-1em}
    \caption{ Sensitivity study for $\nu$ and $\gamma$ on MovieLens Dataset. }
       \label{fig:ablation2}
\vspace{-0.1in}
\end{figure}

\textbf{Study for $\nu$ and $\gamma$}. 
Figure \ref{fig:ablation2} illustrates the performance variation of \sysn concerning the parameters $\nu$ and $\gamma$. For the sake of discussion, we will focus on $\nu$ but note that $\gamma$ plays a similar role in terms of controlling clustering.
When $\nu$ is set to a value like 1.1, the exploration range of clusters becomes very narrow. In this case, the inferred cluster size in each round, $|\widehat{\caln}{u_t}(\bx_{t,i})|$, tends to be small. This means that the inferred cluster $\widehat{\caln}{u_t}(\bx_{t,i})$ is more likely to consist of true members of $u_t$'s relative cluster. However, there is a drawback regarding this narrow exploration range: it might result in missing out on potential cluster members in the initial phases of learning.
On the other hand,
setting $\nu$ to a larger value, like $\nu=5$, widens the exploration range of clusters. This means that there are more opportunities to include a larger number of members in the inferred cluster. However, continuously increasing $\nu$ does not necessarily lead to improved performances, because excessively large values of $\nu$ might result in inferred clusters that include non-collaborative users and clustering noise. Therefore, in practice, we recommend to set $\nu$ to a relatively large number (e.g., $\nu = 5$) that strikes a balance between the exploration and exploitation.

\begin{figure}[ht] 
\vspace{-0.1in}
    \includegraphics[width=0.7\columnwidth]{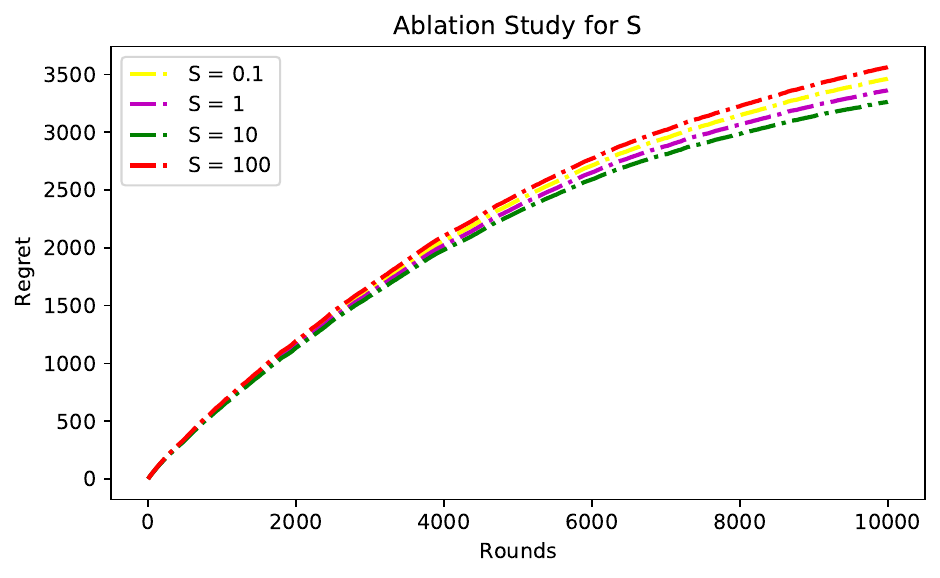}
    \centering
     \vspace{-1em}
    \caption{ Sensitivity study for $S$ on Mnist Dataset. }
       \label{fig:ablation_alpha}
\vspace{-0.1in}
\end{figure}

\textbf{Study for $S$}. 
Figure \ref{fig:ablation_alpha} provides insight into the sensitivity of \sysn concerning the parameter $S$ in Algorithm \ref{alg:main}. It is evident that \sysn exhibits robust performance across a range of values for $S$. This robustness can be attributed to the strong discriminability of the meta-learner and the derived upper bound.
Even with varying $S$ values, the relative order of arms ranked by \sysn experiences only slightly changes. This consistency in arm rankings demonstrates that \sysn is capable of maintaining the robust performance, which in turn reduces the need for extensive hyperparameter tuning.

\section{Conclusion} \label{sec:conclusion}

In this paper, we study the Cluster of Neural Bandits problem to incorporate correlation in bandits with generic reward assumptions. Then, we propose a novel algorithm, \sysn, to solve this problem, where a meta-learner is assigned to represent and rapidly adapt to dynamic clusters, along with an informative UCB-type exploration strategy. 
Moreover, we provide the instance-dependent regret analysis for \sysn.
In the end, to demonstrate the effectiveness of \sysn, we conduct extensive experiments to evaluate its empirical performance against strong baselines on recommendation and classification datasets.

\section*{Acknowledgement}
This work is supported by National Science Foundation under Award No. IIS-2002540, and Agriculture and Food Research Initiative (AFRI) grant no. 2020-67021-32799/project accession no.1024178 from the USDA National Institute of Food and Agriculture. The views and conclusions are those of the authors and should not be interpreted as representing the official policies of the funding agencies or the government.

\bibliographystyle{abbrvnat}
\balance
\bibliography{ref}


\appendix
\onecolumn

\section{Proof Details of Theorem \ref{theorem:main}} \label{sec:proofs}

Our proof technique is different from related works. \citep{2014onlinecluster,2016collaborative,gentile2017context,2019improved,ban2021local} are built on the classic linear bandit framework and \citep{zhou2020neural, zhang2020neural, kassraie2022neural} utilize the kernel-based analysis in the NTK regime. 
In contrast, we use the generalization bound of user-learner to bound the error incurred in each round and bridge meta-learner with user-learner by bounding their distance, which leads to our final regret bound.
Specifically, we decompose the regret of $T$ rounds into three key terms (Eq. (\ref{eq:decomponents})), where the first term is the error induced by user learner $\theta^u $, the second term is the distance between user learner and meta learner, and the third term is the error induced by the meta learner $\Theta$. 
Then, Lemma \ref{lemma:userboundasttheta} provides an upper bound for the first term. Lemma \ref{lemma:userboundasttheta} is an extension of Lemma \ref{lemma:genesingle}, which is the key to removing the input dimension. Lemma \ref{lemma:genesingle} has two terms with the complexity $\calo(\sqrt{T})$, where the first term is the training error induced by a class of functions around initialization, the second term is the deviation induced by concentration inequality for $f(\cdot; \theta^u)$. Lemma \ref{lemma:metauserdistance} bounds the distance between user-learner and meta-learner. 
Lemma \ref{lemma:ucb} bounds the error induced by the meta learner using triangle inequality bridged by the user learner.
Bounding the three terms in Eq. (\ref{eq:decomponents}) completes the proof.

We first show the lemmas for the analysis of user-learner in Section \ref{proof:banditlearner}, the lemmas for meta-learner in Section \ref{proof:metalearner}, the lemma to bridge bandit-learner and meta-learner in Section \ref{proof:bridge}, and the lemmas for the main workflow in Section \ref{sec:mainproof}.

\subsection{Analysis for user-learner}\label{proof:banditlearner}

Following \cite{allen2019convergence, cao2019generalization}, given an instance $\bx \in \bbe^d$ with $\|\bx\|_2 = 1$, we define the outputs of hidden layers of the neural network (Eq. (\ref{eq:structure})):
\[
\bh_0 = \bx,  \bh_l = \sigma(\bw_l \bh_{l-1}), l \in [L-1].
\]
Then, we define the binary diagonal matrix functioning as ReLU:
\[
\bD_l = \text{diag}( \mathbbm{1}\{(\bw_l \bh_{l-1})_1 \}, \dots, \mathbbm{1}\{(\bw_l \bh_{l-1})_m \} ), l \in [L-1].
\]
Accordingly, given an input $\bx$, the neural network (Eq. (\ref{eq:structure})) is represented by 
\[
f(\bx_t; \theta \ \text{or} \ \Theta) = \bw_L (\prod_{l=1}^{L-1} \bD_l \bw_l) \bx,
\]
and
\[
\nabla_{\bw_l}f  = 
\begin{cases}
[\bh_{l-1}\bw_L (\prod_{\tau=l+1}^{L-1} \bD_\tau \bw_\tau)]^\top, l \in [L-1] \\
\bh_{L-1}^\top,   l = L .
\end{cases}
\]
Given a reward $r \in [0, 1]$, define $\call(\theta) = (f(\bx; \theta) - r)^2/2$.
Then, we have the following auxiliary lemmas.

\begin{lemma} \label{lemma:xi1}
 Suppose $m, \eta_1, \eta_2$ satisfy the conditions in Theorem \ref{theorem:main}.  With probability at least $1 -  \calo(TKL)\cdot \exp(-\Omega(m \omega^{2/3}L))$ over the random initialization, for all $t \in [T], i \in [k] $,  $\theta$ (or $\Theta$) satisfying $ \|\theta - \theta_0\|_2 \leq \omega$ with $ \omega \leq \calo(L^{-9/2} [\log m]^{-3})$, it holds uniformly that
\[
\begin{aligned}
&(1), | f(\bx; \theta)| \leq  \calo(1). \\
&(2), \| \nabla_{\theta}    f(\bx; \theta) \|_2 \leq \calo(\sqrt{L}). \\
&(3), \|  \nabla_{\theta} \call(\theta)  \|_2 \leq  \calo(\sqrt{L})
\end{aligned}
\]
\end{lemma}

\begin{proof}
(1) is a simply application of Cauchy–Schwarz inequality.
\[
\begin{aligned}
| f(\bx; \theta)| & = |  \bw_L (\prod_{l=1}^{L-1} \bD_l \bw_l) \bx  |  \leq \underbrace{\|  \bw_L (\prod_{l=1}^{L-1} \bD_l \bw_l)  \|_2}_{I_1} \|\bx \|_2   \leq \calo (1)
\end{aligned}
\]
where $I_1$ is based on the Lemma B.2 in \cite{cao2019generalization}: $I_1 \leq \calo(1)$, and $\|\bx\|_2 = 1$.
For (2), it holds uniformly that
\[
 \| \nabla_{\theta}    f(\bx; \theta) \|_2  = \| \text{vec}( \nabla_{\bw_1}f )^\top,  \dots, \text{vec}( \nabla_{\bw_L}f )^\top\|_2 
 \leq \calo(\sqrt{L})
\]
where $\| \nabla_{\bw_1}f\|_F \leq \calo(1) $ is an application of Lemma B.3 in \cite{cao2019generalization} by removing $\sqrt{m}$.

For (3), we have  $\| \nabla_{\theta} \call(\theta)\|_2 \leq | \call' | \cdot \|   \nabla_\theta  f(\bx; \theta) \|_2 \leq \calo(\sqrt{L})$.
\end{proof}

\begin{lemma} \label{lemma:functionntkbound}
Suppose $m, \eta_1, \eta_2$ satisfy the conditions in Theorem \ref{theorem:main}.  With probability at least $1 -  \calo(TKL)\cdot \exp(-\Omega(m \omega^{2/3}L))$, for all $t \in [T], i \in [k] $, $\theta, \theta'$ (or $\Theta, \Theta'$ ) satisfying $ \|\theta - \theta_0\|_2,  \|\theta' - \theta_0\|_2 \leq \omega$ with $ \omega \leq \calo(L^{-9/2} [\log m]^{-3})$, it holds uniformly that
\[
| f(\bx; \theta) - f(\bx; \theta') -  \langle  \triangledown_{\theta'} f(\bx; \theta'), \theta - \theta'    \rangle    | \leq \mathcal{O} (\omega^{1/3}L^2 \sqrt{ \log m}) \|\theta - \theta'\|_2.
\]
\end{lemma}

\begin{proof}
Based on Lemma 4.1 \cite{cao2019generalization}, it holds uniformly that
\[
|\sqrt{m} f(\bx; \theta) - \sqrt{m} f(\bx; \theta') -  \langle \sqrt{m}  \triangledown_{\theta'}f(\bx; \theta'), \theta - \theta'    \rangle    | \leq \mathcal{O} (\omega^{1/3}L^2 \sqrt{m \log(m)}) \|\theta - \theta'\|_2,
\]
where $\sqrt{m}$ comes from the different scaling of neural network structure. Removing $\sqrt{m}$ completes the proof.
\end{proof}

\begin{lemma}  \label{lemma:differenceee}
Suppose $m, \eta_1, \eta_2$ satisfy the conditions in Theorem \ref{theorem:main}.  With probability at least $1 -  \calo(TKL)\cdot \exp(-\Omega(m \omega^{2/3}L))$, for all $t \in [T], i \in [k] $,  $\theta, \theta'$ satisfying $ \|\theta - \theta_0\|_2,  \|\theta' - \theta_0\|_2 \leq \omega$ with $ \omega \leq \calo(L^{-9/2} [\log m]^{-3})$, it holds uniformly that
\begin{equation}
\begin{aligned}
 (1) \qquad  & |f(\bx; \theta) - f(\bx; \theta')|  \leq   \calo(  \omega \sqrt{L})  +  \mathcal{O} (\omega^{4/3}L^2 \sqrt{ \log m})
 \end{aligned}
\end{equation}
\end{lemma}

\begin{proof}
Based on Lemma \ref{lemma:functionntkbound}, we have
\[
\begin{aligned}
 & |f(\bx; \theta) - f(\bx; \theta')|  \\
\leq  & |    \langle  \nabla_{\theta'}f(\bx; \theta'), \theta - \theta'    \rangle    | +  \mathcal{O} (\omega^{1/3}L^2 \sqrt{ \log m}) \|\theta - \theta'\|_2 \\
\leq & \| \nabla_{\theta'}f(\bx; \theta') \|_2 \cdot \| \theta - \theta' \|_2 + \mathcal{O} (\omega^{1/3}L^2 \sqrt{ \log m}) \|\theta - \theta'\|_2 \\
\leq  &\calo(\sqrt{L})  \| \theta - \theta' \|_2 +  \mathcal{O} (\omega^{1/3}L^2 \sqrt{ \log m}) \|\theta - \theta'\|_2
\end{aligned}
\]
The proof is completed.
\end{proof}

\begin{lemma} [Almost Convexity]  \label{lemma:convexityofloss} 
Let $\call_t(\theta) = (f(\bx_t; \theta) - r_t)^2/2 $.  Suppose $m, \eta_1, \eta_2$ satisfy the conditions in Theorem \ref{theorem:main}.  For any $\epsilon > 0$,  with probability at least $1- \calo(TKL^2)\exp[-\Omega(m \omega^{2/3} L)] $ over randomness of $\theta_1$, for all $ t \in [T]$, and $\theta, \theta'$
satisfying $\| \theta - \theta_0 \|_2 \leq \omega$ and $\| \theta' - \theta_0 \|_2 \leq \omega$ with $\omega \leq \calo(L^{-6} [\log m]^{-3/2} \epsilon^{3/4})$, it holds uniformly that
\[
 \call_t(\theta')  \geq  \call_t(\theta) + \langle \nabla_{\theta}\call_t(\theta),      \theta' -  \theta     \rangle  -  \epsilon.
\]
\end{lemma}

\begin{proof}
Let $\call_t'$ be the derivative of $\call_t$ with respective to $f(\bx_t; \theta)$. Then, it holds that $ |  \call_t' | \leq \calo(1)$ based on Lemma \ref{lemma:xi1}.
Then, by convexity of $\call_t$, we have
\[
\begin{aligned}
&\call_t(\theta') - \call_t(\theta)  \\
\overset{(a)}{\geq} &  \call_t'  [  f(\bx_t; \theta')  -  f(\bx_t; \theta) ]  \\
\overset{(b)}{\geq}  & \call_t' \langle \nabla f(\bx_t; \theta) ,  \theta' -  \theta\rangle  \\
 & - | \call_t'| \cdot  | f(\bx_t; \theta')  -  f(\bx_t; \theta) -   \langle \nabla f(\bx_t; \theta) ,  \theta' -  \theta\rangle  |         \\
\geq & \langle \nabla_{\theta}\call_t(\theta ), \theta' -  \theta \rangle   -  | \call_t'| \cdot |  f(\bx_t; \theta')  -  f(\bx_t; \theta) -   \langle \nabla f(\bx_t; \theta) ,  \theta' -  \theta\rangle  |  \\
 \overset{(c)}{\geq} &  \langle \nabla_{\theta'}\call_t, \theta' -  \theta \rangle  -   \calo(\omega^{4/3}L^3 \sqrt{\log m}))  \\
 \geq & \langle \nabla_{\theta'}\call_t, \theta' -  \theta \rangle - \epsilon
\end{aligned}
\]
where $(a)$ is due to the convexity of $\call_t$, $(b)$ is an application of triangle inequality,  and $(c)$ is the application of Lemma \ref{lemma:functionntkbound}. 
With the choice of $\omega$, the proof is completed.
\end{proof}

\begin{lemma}[User Trajectory Ball] \label{lemma:usertrajectoryball}
Suppose $m, \eta_1, \eta_2$ satisfy the conditions in Theorem \ref{theorem:main}. 
With probability at least $1- \calo(TKL^2)\exp[-\Omega(m \omega^{2/3} L)] $ over randomness of $\theta_0$, for any $R > 0$, it holds uniformly that
\[
\|\theta_t - \theta_0\|_2 \leq \calo(R/m^{1/4}) \leq \calo(L^{-6} [\log m]^{-3/2} T^{-3/4}), t \in [T].
\]
\end{lemma}

\begin{proof}
Let $\omega \leq  \calo(R/m^{1/4})$.
The proof follows a simple induction. Obviously, $\theta_0$ is in $B(\theta_0, \omega)$.  Suppose that $\theta_1, \theta_2, \dots, \theta_T \in \mathcal{B}(\theta_0^2, \omega)$.
We have, for any $t \in [T]$, 
\begin{align*}
\|\theta_T -\theta_0\|_2 & \leq \sum_{t=1}^T \|\theta_{t} -  \theta_{t-1} \|_2 \leq \sum_{t=1}^T \eta \|\nabla \call_t(\Theta_t)\| \leq  \sum_{t=1}^T \eta \sqrt{L} \\
&  =  \calo(TR^2 \sqrt{L} /\sqrt{m}) \leq \calo(R/m^{1/4})
\end{align*}
Then, by the choice of $m$, the proof is complete.
\end{proof}

\begin{lemma} [Instance-dependent Loss Bound] \label{lemma:instancelossbound}
Let $\call_t(\theta) = (f(\bx_t; \theta) - r_t)^2/2$. 
 Suppose $m, \eta_1, \eta_2$ satisfy the conditions in Theorem \ref{theorem:main}. 
With probability at least $1- \calo(TKL^2)\exp[-\Omega(m \omega^{2/3} L)] $ over randomness of $\theta_0$, given any $R > 0$  it holds that
\begin{equation}
\sum_{t=1}^T \call_t(\theta_t)  \leq  \sum_{t=1}^T  \call_t(\theta^\ast) +  \calo(1) +  \frac{TLR^2}{\sqrt{m}}.
\end{equation}
where $\theta^\ast =  \arg \inf_{\theta' \in B(\theta_0, R)}  \sum_{t=1}^T  \call_t(\theta')$.
\end{lemma}

\begin{proof}
In round $t$, based on Lemma \ref{lemma:usertrajectoryball}, 
for any $t \in [T]$,   $\|\theta_t - \theta_0\|_2 \leq O(R/m^{1/4})$, which satisfies the conditions in Lemma \ref{lemma:convexityofloss}. Then, based on \ref{lemma:convexityofloss}, for any $\epsilon > 0$ and all $\theta' \in B(\theta_1, R)$, it holds uniformly

\begin{align*}
\call_t(\theta_t) - \call_t(\theta')  
\leq &  \langle \nabla \call_t(\theta_t),  \theta_t - \theta' \rangle   + \epsilon .
\end{align*}
Therefore,  for all $t \in [T], \theta' \in B(\theta_1, R)$,  it holds uniformly
\begin{align*}
    \call_t(\theta_t) - \call_t(\theta')  \overset{(a)}{\leq} &  \frac{  \langle  \theta_t - \theta_{t+1} , \theta_t - \theta'\rangle }{\eta}  + \epsilon ~\\
   \overset{(b)}{ = }  & \frac{ \| \theta_t - \theta' \|_2^2 + \|\theta_t - \theta_{t+1}\|_2^2 - \| \theta_{t+1} - \theta'\|_2^2 }{2\eta}  + \epsilon ~\\
\overset{(c)}{\leq}& \frac{ \|\theta_t  - \theta'\|_2^2 - \|\theta_{t+1} - \theta'\|_2^2  }{2 \eta}     + O(L\eta) + \epsilon\\
\end{align*}
where $(a)$ is because of the definition of gradient descent, $(b)$ is due to the fact $2 \langle A, B \rangle = \|A \|^2_F + \|B\|_F^2 - \|A  - B \|_F^2$, 
$(c)$ is by $ \|\theta_t - \theta_{t+1}\|^2_2 = \| \eta \nabla_{\theta} \call_t(\theta_t)\|^2_2  \leq \calo(\eta^2L)$.

Then,  for $T$ rounds,  we have
\begin{align*}
   & \sum_{t=1}^T \call_t(\theta_t) -  \sum_{t=1}^T  \call_t(\theta')  \\ 
   \overset{(a)}{\leq} & \frac{\|\theta_1 - \theta'\|_2^2}{2 \eta} + \sum_{t =2}^T \|\theta_t - \theta' \|_2^2 ( \frac{1}{2 \eta} - \frac{1}{2 \eta}    )   + \sum_{t=1}^T L \eta + T \epsilon    \\
   \leq & \frac{\|\theta_1 - \theta'\|_2^2}{2 \eta} + \sum_{t=1}^T L \eta + T \epsilon    \\
   \leq & \calo(\frac{R^2}{ \sqrt{m} \eta}) + \sum_{t=1}^T L \eta + T \epsilon  \\
   \overset{(a)}{\leq} & \calo(1) +  \frac{TLR^2}{\sqrt{m}},
   \end{align*}
where $(a)$ is by simply discarding the last term and $(b)$ is by $\eta = \frac{R^2}{\sqrt{m}}$ and replacing $\epsilon$ with $\calo(1/T)$.  

The proof is completed.
\end{proof}

\begin{lemma} \label{lemma:genesingle}
For any $\delta \in (0, 1),  R > 0 $, and $m , \eta_1, \eta_2$ satisfy the conditions in Theorem \ref{theorem:main}. In a round $\tau$ where $u \in N$ is serving user, let $x_\tau$ be the selected arm and $r_\tau$ is the corresponding received reward.
Then, with probability at least $1 -\delta$ over the randomness of initialization, the cumulative regret induced by $u$ up to round $T$ is upper bounded by:
\[
\begin{aligned}
 & \frac{1}{\mu_T^u} \sum_{ (\bx_\tau, r_\tau ) \in \calt_{T}^u }  \underset{r_\tau | \bx_\tau}{\bbe} [ |f(\bx_\tau; \theta^u_{\tau-1}) - r_\tau|] \\
\leq &  \sqrt{ \frac{S^{\ast}_T(u) + \calo(1) }{\mu_T^u}} + \calo \left(\sqrt{\frac{2 \log( \calo (1) /\delta)}{\mu^u_T}} \right) .
\end{aligned}
\]
where $S^{\ast}_T(u) = \underset{\theta \in B(\theta_0, R)}{\inf} \sum_{ (\bx_\tau, r_\tau) \in  \calt_{T}^u} L_\tau(\theta)$.
\end{lemma}

\begin{proof}
According to Lemma \ref{lemma:xi1}, given any $\|\bx\|_2 = 1, r \leq 1$, for any round $\tau$ in which $u$ is the serving user, we have
\[
| f(\bx; \theta_{\tau-1}^u) - r| \leq  \calo(1).
\]
Then, in a round $\tau$, we define 
\begin{equation}
V_\tau = \underset{ r_\tau | \bx_\tau }{\bbe} [ |f(\bx_\tau; \theta^u_{\tau-1}) - r_\tau| ] - | f(\bx_\tau; \theta^u_{\tau-1}) - r_\tau|,
\end{equation}
where the expectation is taken over $r_\tau$ conditioned on $\bx_\tau$.
Then, we have
\[
\bbe[V_\tau|\mathbf{F}_\tau^u] =  \underset{r_\tau | \bx_\tau }{\bbe} [ |f(\bx_\tau; \theta^u_{\tau-1}) - r_\tau | ] - \bbe[| f(\bx_\tau; \theta^u_{\tau-1}) - r_\tau | \mid \mathbf{F}_\tau^u] = 0
\]
where $\mathbf{F}_\tau^u$ denotes the $\sigma$-algebra generated by $\calt_{\tau-1}^u $. Thus, we have the following form:
\begin{equation}
\frac{1}{\mu_T^u} \sum_{ (\bx_\tau, r_\tau ) \in \calt_{T}^u } V_\tau = \frac{1}{\mu_T^u} \sum_{(\bx_\tau, r_\tau ) \in \calt_{T}^u } \underset{ r_\tau | \bx_\tau }{\bbe} [ |f(\bx_\tau; \theta^u_{\tau-1}) - r_\tau| ] -  \frac{1}{\mu_T^u} \sum_{ (\bx_\tau, r_\tau ) \in \calt_{T}^u  } | f(\bx_\tau; \theta^u_{\tau-1}) - r_\tau |. 
\end{equation}
Because $V_1, \dots, V_{\mu_T^u}$ is the martingale difference sequence, applying Hoeffding-Azuma inequality over  $V_1, \dots, V_{\mu_T^u}$, we have
\begin{equation} \label{eq:singleg1}
\begin{aligned}
&\frac{1}{\mu_T^u} \sum_{ (\bx_\tau, r_\tau ) \in \calt_{T}^u } \underset{r_\tau | \bx_\tau}{\bbe}  [ |f(\bx_\tau; \theta^u_{\tau-1}) - r_\tau|   \mid] \\
\leq & \underbrace{  \frac{1}{\mu_T^u} \sum_{ (\bx_\tau, r_\tau ) \in \calt_{T}^u  } | f(\bx_\tau; \theta^u_{\tau-1}) - r_\tau |}_{I_1} +  \calo \left( \sqrt{\frac{2 \log(1/\delta)}{\mu^u_T}} \right).
\end{aligned}
\end{equation}
For $I_1$, based on Lemma \ref{lemma:usertrajectoryball} and Lemma \ref{lemma:instancelossbound} , we have 
\begin{equation}\label{eq:singleg2}
\begin{aligned}
  & \frac{1}{\mu_T^u} \sum_{ (\bx_\tau, r_\tau ) \in \calt_{T}^u } | f(\bx_\tau; \theta^u_{\tau-1}) - r_\tau |  \\
 \leq & \sqrt{\frac{ \sum_{ (\bx_\tau, r_\tau ) \in \calt_{T}^u } ( f(\bx_\tau; \theta^u_{\tau-1}) - r_\tau )^2}{ \mu_T^u} } \\
 \overset{(a)}{\leq}  &     \sqrt{ \frac{S^{\ast}_T(u)  + \calo(1) }{\mu_T^u}}.
\end{aligned}
\end{equation}
where $(a)$ is an application of Lemma \ref{lemma:instancelossbound}. Combining Eq.(\ref{eq:singleg1}) and Eq.(\ref{eq:singleg2}) and applying the union bound completes the proof. 
\end{proof}

\begin{lemma} \label{lemma:utroundsregret}
For any $\delta \in (0, 1),  R > 0 $, and $m , \eta_1, \eta_2$ satisfy the conditions in Theorem \ref{theorem:main}.
Suppose $ \widehat{\caln}_{u_t}(\bx_t) =  \caln_{u_t}(\bx_t), \forall t \in [T]$.
After $T$ rounds, with probability $1-\delta$ over the random initialization, the cumulative error induced by the bandit-learners is upper bounded by
\[
\begin{aligned}
& \sum_{t=1}^T  \frac{1}{|\caln_{u_t}(\bx_t) |} \sum_{u_{t,i}  \in \caln_{u_t}(\bx_t) }   \underset{  r_t | \bx_t }{\bbe} [ |f(\bx_t; \theta^{u_{t,i}}_{t-1}) - r_t| ] \\
\leq &  \sqrt{ q T \cdot S^{\ast}_{Tk}  \log(\calo (\delta^{-1}))  + \calo(1) }    + \calo(\sqrt{     2 q T \log(\calo (\delta^{-1}))}) ,
\end{aligned}
\]
where  $S^{\ast}_{Tk} = \underset{\theta \in B(\theta_0, R)}{\inf} \sum_{t=1}^{Tk}\call_{t}(\theta)$ .
\end{lemma}

\begin{proof}
Applying Lemma \ref{lemma:genesingle} over all users, we have
\begin{equation}
\begin{aligned}
&  \sum_{t=1}^T    \frac{1}{|\caln_{u_t}(\bx_t) |}   \sum_{u_{t,i}  \in \caln_{u_t}(\bx_t) }  \underset{  r_t | \bx_t }{\bbe} [ |f(\bx_t; \theta^{u_{t, i}}_{t-1}) - r_t|] \\
\leq &  \sum_{t=1}^T    \frac{1}{ \bbe[ |\caln_{u_t}(\bx_t) |]}   \sum_{u_{t,i}  \in \caln_{u_t}(\bx_t) }  \underset{  r_t | \bx_t }{\bbe} [ |f(\bx_t; \theta^{u_{t, i}}_{t-1}) - r_t|]  + \sqrt{2T \log(\delta^{-1})}\\
= &  \frac{q}{n}    \sum_{u \in N} \sum_{ (\bx_\tau, r_\tau ) \in \calt_{T}^u }  \underset{ r_t | \bx_t }{\bbe} [ |f(\bx_\tau; \theta^u_{t-1}) - r_\tau| ]  + \sqrt{2T \log(\delta^{-1})}  \\
\leq &  \frac{q}{n} \sum_{u \in N}  \Bigg  [  \sqrt{ \mu_T^u \cdot S^{\ast}_T(u)  + \calo(1)    } + \calo( \sqrt{2  \mu^u_T  \log( \calo(1) /\delta)}) \Bigg]
\end{aligned}
\end{equation}
where the first inequality is applying the Hoeffding-Azuma inequality to $\caln_{u_t}(\bx_t)$  and the last inequality is based on Lemma \ref{lemma:genesingle}. Then, based on our update rules for user-learner,  we have the fact
\[
\begin{aligned}
 \frac{q}{n} \sum_{u \in N}  \sqrt{ \mu_T^u } 
  & = \frac{q}{n}   \sum_{u \in N}   \sqrt{ \sum_{t=1}^T  1 \{ u_{t, i} = u\} } \\
  & \leq  \frac{q}{n}   \sum_{u \in N}   \sqrt{ \sum_{t=1}^T \bbe[1 \{ u_{t, i} = u\}]  + \sqrt{2T \log(\delta^{-1})}} \\
  & \leq  \calo( \sqrt{Tq}  + \sqrt{q \sqrt{2T \log(\delta^{-1})}} )
\end{aligned}
\]
where the first inequality is applying the Hoeffding-Azuma inequality again and the last inequality is based on the concavity of the square root function. 
Then, keeping the leading term,  we have
\begin{equation}
    \begin{aligned}
& \sum_{t=1}^T \underset{  r_t | \bx_t }{\bbe} [ |f(\bx_t; \theta^{u_t}_{t-1}) - r_t|] \\
\leq &   \sqrt{ q T \cdot S^{\ast}_{Tk}  \log(\calo (\delta^{-1}))  + \calo(1) }    + \calo(\sqrt{     2 q T \log(\calo (\delta^{-1}))}) .   
\end{aligned}
\end{equation}
where $\calo(1)$ is because of the choice of $m$ and $ S^{\ast}_T(u) \leq S^{\ast}_{Tk}$ .
The proof is completed.
\end{proof}

\begin{corollary} \label{lemma:genesingleast}
For any $\delta \in (0, 1),  R > 0 $, and $m , \eta_1, \eta_2$ satisfy the conditions in Theorem \ref{theorem:main}.
In a round $\tau$ where $u \in N$ is the serving user, let $x_\tau^\ast$ be the arm selected according to Bayes-optimal policy $\pi^\ast$: 
\[
x_\tau^\ast = \arg \max_{\bx_{\tau,i}, i \in [k]} h_{u} (\bx_{\tau, i}),  
\]
and $r_\tau^\ast$ is the corresponding reward.
Then, with probability at least $1 -\delta$ over the randomness of initialization, after $t \in [T]$ rounds, the cumulative regret induced by $u$ with policy $\pi^\ast$ is upper bounded by: 
\[
\begin{aligned}
 &        \frac{1}{\mu_t^u} \sum_{ (\bx_\tau^\ast, r_\tau^\ast ) \in \calt_{t}^{u, \ast} }     \underset{  r_\tau^\ast | \bx_\tau^\ast  }{\bbe}  [ |f(\bx_\tau^\ast; \theta^{u, \ast}_{\tau-1}) - r_\tau^\ast| \mid \pi^\ast]  \\
\leq &   \sqrt{ \frac{ S^{\ast}_T (u) + \calo(1) }{\mu_t^u}} + \calo \left(\sqrt{\frac{2 \log( \calo (1) /\delta)}{\mu^u_t}} \right) .
\end{aligned}
\]
where  $S^{\ast}_T (u) = \underset{\theta \in B(\theta_0, R)}{\inf} \sum_{ (\bx_\tau^\ast, r_\tau^\ast) \in  \calt_{t}^{u, \ast}} L_\tau(\theta)$, and $\calt_{t}^{u, \ast} $ are stored Bayes-optimal pairs up to round $t$ for $u$, and $\theta^{u, \ast}_{\tau-1}$ are the parameters trained on  $\calt_{\tau-1}^{u, \ast}$ according to SGD\_User in round $\tau-1$.
\end{corollary}

\begin{proof}
This proof is analogous to Lemma \ref{lemma:genesingle}.
In a round $\tau $ where $u$ is the serving user,  we define 
\begin{equation}
V_\tau =  \underset{ r_\tau^\ast | \bx_\tau^\ast }{\bbe} [ |f(\bx_\tau^\ast; \theta^{u, \ast}_{\tau-1}) - r_\tau^\ast| ] - | f(\bx_{\tau}^\ast; \theta^{u, \ast}_{\tau-1}) - r_\tau^\ast|.
\end{equation}
where the expectation is taken over $ r_\tau^\ast$ conditioned on  $\bx_\tau^\ast$ .
Then, we have
\[
\bbe[V_\tau|\mathbf{F}_\tau] =   \underset{ r_\tau^\ast | \bx_\tau^\ast }{\bbe} [ |f(\bx_\tau^\ast; \theta^{u, \ast}_{\tau-1}) - r_{\tau, \ast}| ] - \bbe[| f(\bx_{\tau}^\ast; \theta^{u, \ast }_{\tau-1}) - r_\tau^\ast| \mid \mathbf{F}_\tau] = 0.
\]
Therefore,  $V_1, \dots, V_{\mu_t^u}$ is the martingale difference sequence. Then, following the same procedure of Lemma \ref{lemma:genesingle}.
\end{proof}

\begin{corollary} \label{lemma:userboundasttheta}
For any $\delta \in (0, 1),  R > 0 $, and $m , \eta_1, \eta_2$ satisfy the conditions in Theorem \ref{theorem:main}.
In round $t \in [T]$, given $u \in N$,
let 
\[
x_t^\ast = \arg \max_{\bx_{t,i}, i \in [k]} h_{u} (\bx_{t, i})  
\]
the  Bayes-optimal arm for $u$ and $r_t^\ast$ is the corresponding reward.
Then, with probability at least $1 -\delta$ over the random initialization, after $T$ rounds, with probability $1-\delta$ over the random initialization, the cumulative error induced by the bandit-learners is upper bounded by:
\[
\begin{aligned}
& \sum_{t=1}^T \underset{  r_t^\ast \mid \bx_t^\ast }{\bbe} [ |f(\bx_t^\ast; \theta^{u_t, \ast}_{t-1}) - r_t^\ast|] \\
\leq &  \sqrt{  q T \cdot S^{\ast}_{Tk}  + \calo(1) }    + \calo(\sqrt{   2 qT \log(\calo (1) /\delta)}).
\end{aligned}
\]
where  the expectation is taken over $r_\tau^\ast$ conditioned on $\bx_\tau^\ast$ and $\theta^{u_t, \ast}_{t-1}$ are the parameters trained on  $\calt_{t-1}^{u_t, \ast}$ according to SGD in round $t-1$.
\end{corollary}

\begin{proof}
The proof is analogous to Lemma \ref{lemma:utroundsregret}.
\end{proof}

\subsection{Analysis for Meta-learner} \label{proof:metalearner}

\begin{lemma}[Meta Trajectory Ball] \label{lemma:metatrajectoryball}
Suppose $m, \eta_1, \eta_2$ satisfy the conditions in Theorem \ref{theorem:main}. 
With probability at least $1- \calo(TKL^2)\exp[-\Omega(m \omega^{2/3} L)] $ over randomness of $\theta_0$, for any $R > 0$, it holds uniformly that
\[
\|\Theta_t - \Theta_0\|_2 \leq \calo(R/m^{1/4}), t \in [T].
\]
\end{lemma}

\begin{proof}
First, for all $t \in [T]$,  we have
\[
\| \nabla_{\Theta_{t-1}} \call_{t-1}(\widehat{\caln}) \|_2  = \frac{1}{|\widehat{\caln}|} \| \sum_{u \in \widehat{\caln}} ( f(\bx; \Theta_{t-1}) - r) \cdot \nabla_\Theta  f(\bx; \Theta_{t-1})  \|_2
 \leq  \frac{1}{|\widehat{\caln}|} \underbrace{\sum_{u \in \widehat{\caln}}   |f(\bx; \Theta_{t-1}) - r|}_{I_1}  \cdot  \underbrace{\|\nabla_\Theta  f(\bx; \Theta_{t-1})\|_2}_{I_2} 
 \leq \calo (\sqrt{L} ) 
\]
where $I_1  \leq \calo(n)$ because $ f(\bx; \Theta), r \leq \calo(1)$ according to Lemma \ref{lemma:xi1} and $I_2 \leq \calo(\sqrt{L})$. 

The proof follows a simple induction. Obviously, $\Theta_0$ is in $B(\theta_0, \omega)$.  Suppose that $\Theta_1, \Theta_2, \dots, \Theta_T \in B(\theta_0, \omega)$.
We have, for any $t \in [T]$, it holds uniformly 
\begin{align*}
\|\Theta_T -\Theta_0\|_2  & \leq \sum_{t=1}^T \|\Theta_{t} -  \Theta_{t-1} \|_2 \leq \sum_{t=1}^T \eta \| \nabla_{\Theta_{t-1}} \call_{t-1}(\widehat{\caln})\|_2 \leq   T  \eta \calo(\sqrt{L}) \\
&  = \calo(  T R^2 /\sqrt{m}) \leq \calo(1/m^{1/4})
\end{align*}
when $m \geq \Omega(T^4 R^8)$.
The proof is completed.
\end{proof}

\begin{lemma} \label{lemma:theorem5allenzhu}
Suppose $m, \eta_1, \eta_2$ satisfy the conditions in Theorem \ref{theorem:main}. 
With probability at least $1- \calo(TKL^2)\exp[-\Omega(m \omega^{2/3} L)] $ over randomness of $\Theta_0$, for all $t \in [T], i \in [K]$, $\Theta$ satisfying $ \| \Theta  - \Theta_0 \|_2 \leq \omega $ with $\omega \leq \calo(L^{-9/2} [\log m]^{-3})$, it holds uniformly:
\[
\| \nabla_\Theta f(\bx_{t,i}; \Theta) -    \nabla_\Theta f(\bx_{t,i}; \Theta_0) \| \leq \calo (\omega L^3   \sqrt{\log m}) \|   \nabla_\Theta f(\bx_{t,i}; \Theta_0)    \|_2 
\]
\end{lemma}

\begin{proof}
This is an application of Theorem 5 \cite{allen2019convergence}. It holds uniformly that
\[
\| \sqrt{m} \nabla_\Theta f(\bx_{t,i}; \Theta) -    \sqrt{m} \nabla_\Theta f(\bx_{t,i}; \Theta_0) \| \leq \calo (\omega L^3  \sqrt{ \log m}) \|  \sqrt{m}  \nabla_\Theta f(\bx_{t,i}; \Theta_0)    \|_2 
\]
where the scale factor $\sqrt{m}$ is because the last layer of neural network in \cite{allen2019convergence} is initialized based on $N(0 ,1)$ while our last layer is initialized from $N(0 ,1/m)$.
\end{proof}

\subsection{ Bridge Meta-learner and User-learner} \label{proof:bridge}

For brevity, we use $ g(\bx_t; \theta_{t-1}^{u})$ to represent the gradient $\nabla_{\theta} f(\bx_t; \theta_{t-1}^{u})$.

\begin{lemma}  \label{lemma:metauserdistance}
Suppose $m, \eta_1, \eta_2$ satisfy the conditions in Theorem \ref{theorem:main}. 
With probability at least $1- \calo(nTKL^2)\exp[-\Omega(m \omega^{2/3} L)] $ over randomness of $\Theta_0$, for all $t \in [T]$, 
any $u \in N$ and $\Theta_t$ returned by Algorithm \ref{alg:meta}, for any $\|\bx_t\|_2 = 1$, it holds uniformly for Algorithms \ref{alg:main}-\ref{alg:meta} that
\begin{equation}
\begin{aligned}
| f(\bx_t; \theta_{t-1}^{u}) -  f(\bx_t; \Theta_{t}) |   \leq 
\frac{R\| \nabla_{\Theta} f(\bx_t; \Theta_{t}) - \nabla_{\theta} f(\bx_t; \theta_{0}^{u})\|_2}{m^{1/4}}  + Z,
\end{aligned}
\end{equation}
where 
\[
Z  =   \frac{\mathcal{O} (RL^2 \sqrt{\log m} ) }{m^{1/3}} +      \frac{\calo ( L^{7/3}R^2  \sqrt{\log m})}{m^{1/2}}  +  \frac{\mathcal{O}(2 R \sqrt{L}) }{ m^{1/4}}. \\ 
\]
\end{lemma}

\begin{proof}

First, we have
\begin{equation} \label{eq8}
\begin{aligned}
  | f(\bx_t; \theta_{t-1}^{u}) -  f(\bx_t; \Theta_{t}) | \leq &   \underbrace{|  f(\bx_t; \theta_{t-1}^{u})  -   \langle g(\bx_t; \theta_{t-1}^{u}), \theta_{t-1}^u - \theta_0^u \rangle  - f(\bx_t; \theta_0^u)  | }_{I_1}  \\
& +        \underbrace{ |  \langle g(\bx_t; \theta_{t-1}^{u}), \theta_{t-1}^u - \theta_0^u \rangle  + f(\bx_t; \theta_0^u)    -   f(\bx_t; \Theta_{t})| }_{I_2},
\end{aligned}
\end{equation}
where the inequality is using Triangle inequality.
$I_1$ is an application of Lemma \ref{lemma:functionntkbound}:
\[
I_1   \leq \mathcal{O} (w^{4/3}L^2 \sqrt{\log(m)}), 
\]
where the second equality is based on the Lemma \ref{lemma:metatrajectoryball}. 

For $I_2$, we have 
\[
\begin{aligned}
 &|  \langle g(\bx_t; \theta_{t-1}^{u}), \theta_{t-1}^u - \theta_0^u \rangle   + f(\bx_t; \theta_0^u)  -   f(\bx_t; \Theta_{t})|  \\
\leq  &  |\langle g(\bx_t; \theta_{t-1}^{u}), \theta_{t-1}^u - \theta_0^u \rangle - \langle g(\bx_t; \Theta_{t}), \Theta_{t} - \Theta_0 \rangle   |\\
 & + |    \langle g(\bx_t; \Theta_{t}), \Theta_{t} - \Theta_0 \rangle  + f(\bx_t; \theta_0^u)       -   f(\bx_t; \Theta_{t})| \\
\leq &  \underbrace{ |\langle g(\bx_t; \theta_{t-1}^{u}), \theta_{t-1}^u - \theta_0^u \rangle - \langle g(\bx_t; \theta_{0}^{u}), \Theta_{t} - \Theta_0 \rangle   | }_{I_3} \\
 & + \underbrace{ |  \langle g(\bx_t; \theta_{0}^{u}), \Theta_{t} - \Theta_0 \rangle       -  \langle g(\bx_t; \Theta_{t}), \Theta_{t} - \Theta_0 \rangle|}_{I_4}  \\
 &+  \underbrace{ |    \langle g(\bx_t; \Theta_{t}), \Theta_{t} - \Theta_0 \rangle + f(\bx_t; \theta_0^u)  -    f(\bx_t; \Theta_{t})| }_{I_5}
\end{aligned}
\]
where the inequalities use Triangle inequality. For $I_3$, we have
\[
\begin{aligned}
 & |\langle g(\bx_t; \theta_{t-1}^{u}), \theta_{t-1}^u - \theta_0^u \rangle - \langle g(\bx_t; \theta_{0}^{u}), \Theta_{t} - \Theta_0 \rangle   | \\
 \leq & |\langle g(\bx_t; \theta_{t-1}^{u}), \theta_{t-1}^u - \theta_0^u \rangle - \langle g(\bx_t; \theta_{0}^{u}), \theta_{t-1}^u - \theta_0^u \rangle| +  | \langle g(\bx_t; \theta_{0}^{u}), \theta_{t-1}^u - \theta_0^u \rangle  - \langle g(\bx_t; \theta_{0}^{u}), \Theta_{t} - \Theta_0 \rangle | \\
 \leq & \underbrace{\| g(\bx_t; \theta_{t-1}^{u}) -  g(\bx_t; \theta_{0}^{u})  \|_2 \cdot \| \theta_{t-1}^u - \theta_0^u\|_2}_{M_1} +     \underbrace{\|g(\bx_t; \theta_{0}^{u})\|_2 \cdot \|\theta_{t-1}^u - \theta_0^u - ( \Theta_{t} - \Theta_0   ) \|_2}_{M_2}
\end{aligned}
\]
For $M_1$, we have
\begin{equation}\label{eq11}
\begin{aligned}
M_1 & \overset{(a)}{\leq}   \omega    \cdot  \| g(\bx_t; \theta_{t-1}^{u}) -  g(\bx_t; \theta_{0}^{u})  \|_2\\
& \overset{(b)}{\leq} \omega  \cdot \calo (\omega L^3  \sqrt{\log m}) \|  g(\bx_t; \theta_{0}^{u})    \|_2  \\
& =  \calo (\omega^2 L^{7/3}  \sqrt{\log m})
\end{aligned}
\end{equation}
where $(a)$ is the application of Lemma \ref{lemma:usertrajectoryball} and $(b)$ utilizes Lemma \ref{lemma:theorem5allenzhu} with Lemma \ref{lemma:usertrajectoryball}.

For $M_2$, we have
\begin{equation}
\begin{aligned}
  M_2 
 \leq  \|  g(\bx_t; \Theta_{0}) \|_2 \left( \| \theta_{t-1}^u - \theta_0^u  \|_2   +  \|  \Theta_{t} - \Theta_0  \|_2       \right)
 \overset{(a)}{\leq}      \mathcal{O}(\sqrt{L})          \cdot  ( 2 \omega  ) \\
\end{aligned}
\end{equation}\label{eq12}
where $(a)$ uses Lemma \ref{lemma:xi1}, \ref{lemma:usertrajectoryball}, and \ref{lemma:metatrajectoryball}. 
Thus, we have
\begin{equation} \label{eq13}
I_3 \leq \calo (\omega^2 L^{7/3} \sqrt{\log m} + 2 \omega \sqrt{L}).
\end{equation}

For $I_4$, we have
\begin{equation} \label{eq14}
\begin{aligned}
I_4 =  & |  \langle g(\bx_t; \Theta_{0}), \Theta_{t} - \Theta_0 \rangle       -  \langle g(\bx_t; \Theta_{t}), \Theta_{t} - \Theta_0 \rangle| \\
\overset{(a)}{\leq}& \| g(\bx_t; \Theta_{t}) - g(\bx_t; \Theta_{0})\|_2 \|   \Theta_{t} - \Theta_0    \|_2 \\
\overset{(b)}{\leq} & \omega \cdot  \| g(\bx_t; \Theta_{t}) - g(\bx_t; \Theta_{0})\|_2 
\end{aligned}
\end{equation}
where $(a)$ is because of Cauchy–Schwarz inequality and the last inequality is by Lemma \ref{lemma:metatrajectoryball}.

For $I_5$, we have
\begin{equation}\label{eq15}
I_5 = |    \langle g(\bx_t; \Theta_{t}), \Theta_{t} - \Theta_0 \rangle + f(\bx_t; \Theta_0)  -    f(\bx_t; \Theta_{t})| \\
\overset{(a)}{\leq}  \calo (w^{1/3}L^2 \sqrt{\log(m)}) \|\Theta_t  - \Theta_0\|_2 \overset{(b)}{\leq}   \calo(w^{4/3}L^2 \sqrt{\log(m)}),
\end{equation}
where $(a)$ is an application of Lemma \ref{lemma:functionntkbound} and $(b)$ uses Lemma \ref{lemma:metatrajectoryball}.

Combing Eq.(\ref{eq8}), (\ref{eq13}), (\ref{eq14}), and (\ref{eq15}), we have
\[
\begin{aligned}
& | f(\bx_t; \theta_{t-1}^{u}) -  f(\bx_t; \Theta_{t}) | \\
 \leq & \omega \cdot  \| g(\bx_t; \Theta_{t}) - g(\bx_t; \Theta_{0})\|_2  + \mathcal{O} (w^{4/3}L^2 \sqrt{\log(m)}) +       \calo (\omega^2 L^{7/3}  \sqrt{\log m})  +  \mathcal{O}(2\omega \sqrt{L})
\end{aligned} 
\]
Replacing $\omega$ with $R/m^{1/4}$ completes the proof. 
\end{proof}

\begin{lemma} \label{lemma:ucb}
Suppose $m, \eta_1, \eta_2$ satisfy the conditions in Theorem \ref{theorem:main}. Then, with probability at least $1- \delta$ over the random initialization, for any $\delta \in (0, 1), R> 0$, after $t$ rounds, the error induced by meta-learner is upper bounded by:
\begin{equation}
\begin{aligned}
&    \sum_{t=1}^T  \underset{ r_t \mid \bx_t}{\bbe} \left[ |f(\bx_{t}; \Theta_{t}) - r_{t}|  \mid u_t \right] \\
\leq &
     \sum_{t=1}^T   \frac{R\| g(\bx_t; \Theta_{t}) - g(\bx_t; \theta_{0}^{u_t})\|_2}{m^{1/4}} + 
  \sum_{u \in N}  \mu_T^{u}   \Bigg  [  \calo \left( \frac{\sqrt{S^{\ast}_{Tk}  + \calo(1) } }{\sqrt{2\mu^{u}_{T}}}  \right)+   \sqrt{\frac{  2 \log(  \calo(1) /\delta) }{\mu^{u}_T}} \Bigg].
\end{aligned}
\end{equation}
where the expectation is taken over $r_t$ conditioned on $x_t$.
\end{lemma}

\begin{proof}
\begin{equation}
\begin{aligned}
 &  \sum_{t=1}^T  \underset{ r_t \mid \bx_t}{\bbe} \left[ |f(\bx_{t}; \Theta_{t}) - r_{t} |  | u_t \right] \\
 = &   \sum_{t=1}^T \underset{ r_t \mid \bx_t}{\bbe} [| f(\bx_{t}; \Theta_{t}) -f(\bx_{t}; \theta^{u_t}_{t-1}) + f(\bx_{t}; \theta^{u_t}_{t-1}) - r_{t} | \mid u_t ] \\
 \leq &  \underbrace{\sum_{t=1}^T  | f(\bx_{t}; \Theta_{t}) -f(\bx_{t}; \theta^{u_t}_{t-1}) |}_{I_1} +    \underbrace{  \sum_{t=1}^T  \underset{ r_t \mid \bx_t}{\bbe}[ |f(\bx_{t}; \theta^{u_t}_{t-1}) - r_{t}| \mid u_t]}_{I_2} \\.
\end{aligned}
\end{equation}
For $I_1$, applying Lemma \ref{lemma:metauserdistance}, with probability at least $1 -\delta$, for any $\|\bx_{t} \|_2 =1$,  we have
\begin{equation}
I_1 \leq  \sum_{t=1}^T   \frac{R\| g(\bx_t; \Theta_{t}) - g(\bx_t; \theta_{0}^{u_t})\|_2}{m^{1/4}}  + T \cdot Z
\end{equation}
where we ignore the last term as the result of the choice of $m$  for $Z$.

For $I_2$, based on the Lemma \ref{lemma:genesingle}, with probability at least $1 -\delta$, we have
\begin{equation}
\begin{aligned}
I_2  &\leq \sum_{u \in N}  \Bigg  [  \sqrt{ \mu_T^u \cdot S^{\ast}_{Tk}  + \calo(1)    } + \calo( \sqrt{2  \mu^u_T  \log( \calo(1) /\delta)}) \Bigg]
\end{aligned}\end{equation}
The proof is complete. 
\end{proof}

\subsection{ Main Proof } \label{sec:mainproof}

\theoremmain*

\begin{proof}
  Let $ \bx_t^\ast = \arg \max_{\bx_{t, i} \in \BX_t} h_{u_t}(\bx_{t,i})$ given $\BX_t, u_t$, and let $\Theta^\ast_t $ be corresponding parameters trained by Algorithm \ref{alg:meta} based on $\widehat{\mathcal{N}}_{u_t}(\bx_t^\ast)$.
Then, for the regret of one round $t \in [T]$, we have
\[
\begin{aligned}
& R_t \\
 = & \underset{ r_{t,i} \mid \bx_{t,i}, i \in [k] }{\bbe} \left [ r^\ast_t - r_t \mid u_t   \right] \\
 = &     \underset{ r_{t,i} \mid \bx_{t,i}, i \in [k] }{\bbe} \left [  r^\ast_t  - f(\bx_t^\ast; \theta_{t-1}^{u_t, \ast})   + f(\bx_t^\ast; \theta_{t-1}^{u_t, \ast}) -      r_t  \right]\\
 = & 
  \underset{ r_{t,i} \mid \bx_{t,i}, i \in [k] }{\bbe} \left [  r^\ast_t  - \frac{1}{|\caln_{u_t} (\bx_t^\ast) |}  \sum_{u_{t,i} \in \caln_{u_t} (\bx_t^\ast) } f(\bx_t^\ast; \theta_{t-1}^{u_{t,i}, \ast})  +  \frac{1}{|\caln_{u_t} (\bx_t^\ast) |}  \sum_{u_{t,i} \in \caln_{u_t} (\bx_t^\ast) } f(\bx_t^\ast; \theta_{t-1}^{u_{t,i}, \ast}) -  f(\bx_t^\ast; \Theta_{t}^\ast)   
     + f(\bx_t^\ast; \Theta_{t}^\ast )  - r_t    \right] \\
\leq & \frac{1}{|\caln_{u_t} (\bx_t^\ast) |}  \sum_{u_{t,i} \in \caln_{u_t} (\bx_t^\ast) } \underset{  r^\ast_t \mid \bx_t^\ast   }{\bbe} \left [  r^\ast_t  - f(\bx_t^\ast; \theta_{t-1}^{u_{ti}, \ast})  \right] + \frac{1}{|\caln_{u_t} (\bx_t^\ast) |}  \sum_{u_{t,i} \in \caln_{u_t} (\bx_t^\ast) } | f(\bx_t^\ast; \theta_{t-1}^{u_{t, i}, \ast}) -  f(\bx_t^\ast; \Theta_{t}^\ast)  |    \\
& +  \underset{ r_t \mid \bx_t } {\bbe} \left [   f(\bx_t^\ast; \Theta_{t}^\ast )   - r_t     \right] 
\end{aligned}
\]
where the expectation is taken over $r_{t,i}$ conditioned on $\bx_{t,i}$ for each $i \in [k]$,
$ \theta_{t-1}^{u_t, \ast}$ are intermediate user parameters introduced in Lemma \ref{lemma:genesingleast} trained on Bayes-optimal pairs by Algorithm \ref{alg:main}, e.g., $(\bx_{t-1}^\ast, r_{t-1}^\ast)$, and $\Theta_{t}^\ast $ are meta parameters trained on the group $\widehat{\mathcal{N}}_{u_t}(\bx_t^\ast) $ using Algorithm \ref{alg:meta}.
Then, the cumulative regret of $T$ rounds can be upper bounded by
\begin{equation} \label{eq:decomponents}
\begin{aligned}
& \sum_{t=1}^T R_t  \\
\leq &  \sum_{t=1}^T  \frac{1}{|\caln_{u_t} (\bx_t^\ast) |}  \sum_{u_{t,i} \in \caln_{u_t} (\bx_t^\ast) }   \underset{r_t^\ast \mid \bx_t^\ast  }{\bbe} \left [ | r^\ast_t  - f(\bx_t^\ast; \theta_{t-1}^{u_{t, i}, \ast})  | \right] +  \sum_{t=1}^T \frac{1}{|\caln_{u_t} (\bx_t^\ast) |}  \sum_{u_{t,i} \in \caln_{u_t} (\bx_t^\ast) }   |  f(\bx_t^\ast; \theta_{t-1}^{u_{t, i}, \ast}) -  f(\bx_t^\ast; \Theta_{t}^\ast)   |  +  \sum_{t=1}^T   \underset{ r_t \mid \bx_t  }{\bbe} \left [   f(\bx_t^\ast; \Theta_{t}^\ast )   - r_t     \right] \\
\overset{(a)}{\leq} &  \frac{q}{n} \sum_{u \in N}  \Bigg  [   \calo \left(\sqrt{ 2\mu^{u}_{T} S^{\ast}_{TK} + \calo(1) }  \right)+   \sqrt{  2 \log( \mu^{u}_T  \calo(1) /\delta)}   \Bigg]  +  \sum_{t=1}^T \left[ 
Rm^{-1/4} \cdot  \| g(\bx_t^\ast; \Theta_{t}^\ast) - g(\bx_t^\ast; \theta_{0}^{u_t, \ast})\|_2  +
T \cdot Z \right] +  \sum_{t=1}^T   \underset{r_t \mid \bx_t  }{\bbe} \left [   f(\bx_t^\ast; \Theta_{t}^\ast )   - r_t     \right]  \\
\overset{(b)}{\leq} & \frac{q}{n} \sum_{u \in N}  \Bigg  [   \calo \left(\sqrt{ 2\mu^{u}_{T} S^{\ast}_{TK} + \calo(1) }  \right)+   \sqrt{  2 \log( \mu^{u}_T  \calo(1) /\delta)}   \Bigg]   +  \sum_{t=1}^T \left[ 
Rm^{-1/4} \cdot  \| g(\bx_t; \Theta_{t}) - g(\bx_t; \theta_{0}^{u_t})\|_2  +
T \cdot Z \right] +  \underbrace{\sum_{t=1}^T   \underset{r_t \mid \bx_t  }{\bbe} \left [   f(\bx_t; \Theta_{t} )    - r_t     \right]}_{I_1}  \\
\end{aligned}
\end{equation}
where $(a)$ is the applications of Lemma \ref{lemma:userboundasttheta} and Lemma \ref{lemma:metauserdistance}, and (b) is due to the selection criterion of Algorithm \ref{alg:main} where $\theta_{0}^{u_t} = \theta_{0}^{u_t, \ast}$ according to our initialization.
For $I_1$, based on Lemma \ref{lemma:ucb}, we have
\[
\begin{aligned}
I_1 
\leq &    \frac{q}{n} \sum_{u \in N}  \Bigg  [   \calo \left(\sqrt{ 2\mu^{u}_{T} S^{\ast}_{TK} + \calo(1) }  \right)+   \sqrt{  2 \log( \mu^{u}_T  \calo(1) /\delta)}   \Bigg]   +   \sum_{t=1}^T \left[ 
Rm^{-1/4} \cdot  \| g(\bx_t; \Theta_{t}) - g(\bx_t; \Theta_{0})\|_2  +
T \cdot Z \right]. \\
\end{aligned}
\]
Thus, we have 
\[
\begin{aligned}
 \sum_{t=1}^T R_t  \leq  & 3 \underbrace{  \frac{q}{n} \sum_{u \in N}  \Bigg  [   \calo \left(\sqrt{ 2\mu^{u}_{T} S^{\ast}_{TK} + \calo(1) }  \right)+   \sqrt{  2 \log( \mu^{u}_T  \calo(1) /\delta)}   \Bigg]   }_{I_1}  + \underbrace{ 2 \sum_{t=1}^T \left[ 
Rm^{-1/4} \cdot  \| g(\bx_t; \Theta_{t}) - g(\bx_t; \Theta_{0})\|_2  +
T \cdot Z \right]}_{I_2}
\end{aligned}
\]

Inspired Lemma \ref{lemma:utroundsregret}, it holds that
\[
I_1 \leq  \sqrt{ q T \cdot S^{\ast}_{Tk}  \log(\calo (\delta^{-1}))  + \calo(1) }    + \calo(\sqrt{     2 q T \log(\calo (\delta^{-1}))}),
\]
For $I_2$, based on Lemma \ref{lemma:theorem5allenzhu}, we have
\[
I_2 \leq \frac{\calo ( R \omega L^3 m^{-1/2}  \sqrt{\log m}) \|   \nabla_\Theta f(\bx_{t,i}; \Theta_0)    \|_2  }{m^{1/4}} \leq \calo(1)
\]
where $\omega$ is replaced by $R/m^{1/4}$ and because of the choice of $m$ with $ \|   \nabla_\Theta f(\bx_{t,i}; \Theta_0)    \|_2 \leq \calo(\sqrt{L})$ based on Lemma \ref{lemma:xi1}.
Putting them together, the proof is completed.
\end{proof}

\section{Connections with Neural Tangent Kernel}

\begin{lemma} [Lemma \ref{lemma:upperboundofStk} Restated]
Suppose $m$ satisfies the conditions in Theorem \ref{theorem:main}. With probability at least $1 - \delta$ over the initialization, there exists $\btheta'  \in  B(\theta_0, \widetilde{\Omega}(T^{3/2}))$, such that
\[
\bbe[S^{\ast}_{Tk}] \leq    \sum_{t=1}^{TK} \bbe [ ( r_t -  f(\bx_t; \theta'))^2/2 ] \leq \calo \left(\sqrt{ \widetilde{d} \log(1 + TK) - 2 \log \delta  } + S + 1 \right)^2 \cdot \widetilde{d} \log (1+TK).
\]
\end{lemma}

\begin{proof}
\[
\begin{aligned}
& \bbe [ \sum_{t=1}^{TK} (r_t -  f(\bx_t; \theta') )^2 ] \\
 = &  \sum_{t=1}^{TK} ( h(\bx_t) - f(\bx_t; \theta') )^2  \\
\overset{(a)}{\leq} &   \calo \left ( \sqrt{  \log \left(  \frac{\deter(\mathbf{A}_T)} { \deter(\mathbf{I}) }   \right)   - 2 \log  \delta }  +  S  + 1  \right)^2 \sum_{t=1}^{TK}   \| \gx  \|_{\mathbf{A}_{T}^{-1}}^2 +  2TK \cdot \calo \left(\frac{T^2 L^3 \sqrt{\log m}}{m^{1/3}} \right) \\
 \overset{(b)}{\leq} &  \calo \left(\sqrt{ \widetilde{d} \log(1 + TK) - 2 \log \delta  } + S + 1 \right)^2 \cdot 
 \left( \widetilde{d} \log (1+TK) + 1 \right) + \calo(1),
\end{aligned}
\]
where $(a)$ is based on Lemma \ref{lemma:bounfofsinglethetaprime} and $(b)$ is an application of Lemma 11 in \cite{2011improved} and Lemma \ref{lemma:detazero}, and $\calo(1)$ is induced by the choice of $m$.
By ignoring $\calo(1)$,  The proof is completed.
\end{proof}

\begin{definition} \label{def:ridge}
Given the context vectors $\{\bx_i\}_{i=1}^T$ and the rewards $\{ r_i \}_{i=1}^{T} $, then we define the estimation $\widehat{\theta}_t$ via ridge regression:  
\[
\begin{aligned}
&\mathbf{A}_t = \mathbf{I} + \sum_{i=1}^{t} g(\bx_i; \theta_0) g(\bx_i; \theta_0)^\top \\
&\mathbf{b}_t = \sum_{i=1}^{t} r_i g(\bx_i; \theta_0)  \\
&\widehat{\theta}_t = \mathbf{A}^{-1}_t \mathbf{b}_t 
\end{aligned}
\]
\end{definition}

\begin{lemma}\label{lemma:bounfofsinglethetaprime}
Suppose $m$ satisfies the conditions in Theorem \ref{theorem:main}. With probability at least $1 - \delta$ over the initialization, there exists $\btheta'  \in  B(\theta_0, \widetilde{\Omega}(T^{3/2}))$ for all $t \in [T]$, such that
\[
| \hx  - f(\bx_t; \theta') | \leq  \calo \left ( \sqrt{  \log \left(  \frac{\deter(\mathbf{A}_t)} { \deter(\mathbf{I}) }   \right)   - 2 \log  \delta }  +  S  + 1  \right) \| \gx  \|_{\mathbf{A}_{t}^{-1}} + \calo \left(\frac{T^2 L^3 \sqrt{\log m}}{m^{1/3}} \right)
\]
\end{lemma}

\begin{proof}
Given a set of context vectors $\{\bx\}_{t=1}^{T}$ with the ground-truth function $h$ and a fully-connected neural network $f$, we have
\[
\begin{aligned}
 &\left| \hx  - f(\bx_t; \theta')     \right| \\
\leq &  \left  | \hx  - \langle \gx,  \htheta_t  \rangle \right|  + \left| f(\bx_t; \btheta')  -  \langle \gx, \htheta_t \rangle  \right|
\end{aligned}
\]
where $\htheta_t$ is the estimation of ridge regression from Definition \ref{def:ridge}. Then, based on the Lemma \ref{lemma:existthetastar},
there exists $\theta^\ast \in \mathbf{R}^{P}$ such that $ h(\bx_i) =  \left \langle  g(\bx_i, \btheta_0), \theta^\ast \right \rangle$. Thus, we have
\[
\begin{aligned}
& \ \   \left  | \hx  - \langle \gx, \htheta_t \rangle \right| \\
  = & \left|  \left \langle  g(\bx_i, \btheta_0),   \theta^\ast \right \rangle   -   \left \langle  g(\bx_i, \btheta_0),  \htheta_t \right \rangle \right | \\
\leq   & \calo \left ( \sqrt{  \log \left(  \frac{\deter(\mathbf{A}_t)} { \deter(\mathbf{I}) }   \right)   - 2 \log  \delta }  +  S   \right) \| \gx  \|_{\mathbf{A}_{t}^{-1}}
\end{aligned}
\]
where the final inequality is based on the the Theorem 2 in \cite{2011improved}, with probability at least $1-\delta$, for any $t \in [T]$.

Second, we need to bound 
\[
\begin{aligned}
&\left| f(\bx_t; \theta') -  \langle g(\bx_t; \btheta_0), \htheta_t \rangle  \right| \\
 \leq & \left |  f(\bx_t; \theta') - \langle g(\bx_t; \btheta_0), \btheta' - \btheta_0 \rangle   \right|  \\
 &+    \left|     \langle  g(\bx_t; \btheta_0), \btheta' - \btheta_0 \rangle   - \langle  g(\bx_t; \btheta_0), \htheta_t \rangle \right|    
\end{aligned} 
\] 
To bound the above inequality,  we first bound
\[
\begin{aligned}
 & \left |  f(\bx_t; \theta') - \langle g(\bx_t; \btheta_0), \btheta' - \btheta_0 \rangle   \right| \\
=& \left |  f(\bx_t; \theta') - f(\mathbf{x}_t; \btheta_0)   - \langle g(\bx_t; \btheta_0), \btheta' - \btheta_0 \rangle   \right| \\
\leq  & \calo(\omega^{4/3} L^3 \sqrt{ \log m}) 
\end{aligned}
\] 
where  we  initialize $ f(\mathbf{x}_t; \btheta_0) = 0$ following \cite{zhou2020neural} and the inequality is derived by Lemma \ref{lemma:functionntkbound} with $\omega = \frac{\calo(t^{3/2})}{m^{1/4}}$. 
Next, we need to bound
\[ 
\begin{aligned}
 &|  \langle  g(\bx_t; \btheta_0), \btheta' - \btheta_0 \rangle - \langle  g(\bx_t; \btheta_0), \htheta_t \rangle | \\
 = & |\langle g(\bx_t; \btheta_0) ,     (\btheta' - \btheta_0 -  \htheta_t ) \rangle|  \\ 
\leq & \| g(\bx_t; \btheta_0)\|_{\mathbf{A}_t^{-1}} \cdot  \| \btheta' - \btheta_0  - \htheta_t\|_{\mathbf{A}_t} \\
\leq & \| g(\bx_t; \btheta_0)  \|_{\mathbf{A}_t^{-1}} \cdot  \|{\mathbf{A}_t} \|_2  \cdot  \| \btheta' - \btheta_0  - \htheta_t\|_2. \\
\end{aligned} 
\]
Due to the Lemma \ref{lemma:detazero} and Lemma \ref{lemma:2thetab}, we have
\[
\begin{aligned}
 & \|{\mathbf{A}_t} \|_2 \cdot   \| \theta' - \btheta_0  - \htheta_t\|_2 \leq  (1 + t \mathcal{O}(L))   \cdot \frac{1}{1 + \calo(tL)}  =\calo(1).
 \end{aligned}
\] 
Finally, putting everything together, we have
\[
\begin{aligned}
\left| \hx  - f(\bx_t; \theta')     \right| &\leq \gamma_1  \| g(\bx_t; \btheta_0) \|_{\mathbf{A}_t^{-1}}  + \gamma_2.    \\
\end{aligned}
\]
The proof is completed.
\end{proof}

\begin{definition}
\[
\begin{aligned}
&\mathbf{G}^{(0)} = \left[ g(\bx_1; \theta_0), \dots,   g(\bx_T; \theta_0)\right]  \in \bbr^{p \times T}   \\
&\mathbf{G}_0 = \left[ g(\bx_1; \theta_0), \dots,   g(\bx_{TK}; \theta_0) \right]  \in \bbr^{p \times TK} \\
&\mathbf{r}= (r_1, \cdots, r_T) \in \bbr^T
\end{aligned}
\]
$\mathbf{G}^{(0)}$ and $\mathbf{r}$ are formed by the selected contexts and observed rewards in $T$ rounds, $\mathbf{G}_0$ are formed by all the presented contexts.

Inspired by Lemma B.2 in \cite{zhou2020neural} , with $\eta = m^{-1/4}$ we define the auxiliary sequence following :
\[
\theta_0 = \theta^{(0)}, \ \ \theta^{(j+1)}   = \theta^{(j)} - \eta\left[ \bsg^{(0)} \left( [\bsg^{(0)}]^\top (\theta^{(j)}  - \theta_0) - \bsr \right)  + \lambda (\theta^{(j)} - \theta_0 )  \right] 
\]
\end{definition}

\begin{lemma} \label{lemma:existthetastar}
Suppose $m$ satisfies the conditions in Theorem \ref{theorem:main}. With probability at least $1 - \delta$ over the initialization, for any $t \in [T], i \in [K]$, the result uniformly holds:
\[
h_{u_t}(\bx_{t,i}) = \langle g(\bx_{t,i}; \theta_0), \theta^\ast - \theta_0 \rangle.
\]

\end{lemma}
\begin{proof}
Based on Lemma \ref{lemma:boundgradientandNTK} with proper choice of $\epsilon$, we have
\[
\mathbf{G}^\top_0 \mathbf{G}_0 \succeq  \mathbf{H} - \|   \mathbf{G}^\top_0 \mathbf{G}_0 - \mathbf{H}    \|_F \mathbf{I} \succeq \mathbf{H} -\lambda_0 \mathbf{I}/2 \succeq \mathbf{H}/2 \succeq 0.
\] 
Define $\mathbf{h} = [h_{u_1}(\bx_1), \dots, h_{u_T}(\bx_{TK})]$.
Suppose the singular value decomposition of $\mathbf{G}_0$ is $\mathbf{PAQ}^\top,   \mathbf{P} \in \bbr^{p \times TK},  \mathbf{A} \in \bbr^{TK \times TK},  \mathbf{Q} \in \bbr^{TK \times TK}$, then, $\mathbf{A} \succeq 0$. 
Define  $\theta^\ast = \theta_0 + \mathbf{P} \mathbf{A}^{-1} \mathbf{Q}^\top \mathbf{h}$. Then, we have 
\[
\mathbf{G}^\top_0 (\btheta^\ast - \btheta_0) = \mathbf{QAP}^\top \mathbf{P}\mathbf{A}^{-1} \mathbf{Q}^{\top} \mathbf{h} = \mathbf{h}.
\]
which leads to 
\[
 \sum_{t=1}^{T}  \sum_{i = 1}^K ( h_{u_t}(\bx_{t, i}) - \langle g(\bx_{t, i}; \theta_0), \theta^\ast - \btheta_0 \rangle ) = 0.
\]
Therefore, the result holds:
\begin{equation}
\|     \btheta^\ast - \btheta_0       \|_2^2 = \mathbf{h}^\top \mathbf{QA}^{-2}\mathbf{Q}^\top \mathbf{h} =  \mathbf{h}^\top (\mathbf{G}^\top_0 \mathbf{G}_0)^{-1} \mathbf{h}  \leq  2 \mathbf{h}^\top\mathbf{H}^{-1} \mathbf{h} 
\end{equation}

\end{proof}

\begin{lemma} \label{lemma:2thetab}
There exist $\btheta'  \in  B(\theta_0, \wcalo(T^{3/2}L + \sqrt{T}))$, such that,  with probability at least $1 -\delta$, the results hold:
\[
\begin{aligned}
& (1) \|         \btheta' - \theta_0 \|_2 \leq  \frac{ \wcalo(T^{3/2}L + \sqrt{T })}{m^{1/4}} \\
& (2) \| \theta' - \theta_0 - \widehat{\theta}_t  \|_2  \leq \frac{1}{1 +  \calo(TL)} \\
\end{aligned}
\]
\end{lemma}

\begin{proof}
\normalfont
The sequence of $\theta^{(j)}$ is updates by using gradient descent on the loss function:
\[
\min_{\theta} \mathcal{L}(\theta) = \frac{1}{2}  \|[\bsg^{(0)}]^\top (\theta - \btheta^{(0)} ) - \bsr  \|^2_2 + \frac{m\lambda}{2} \|\theta - \btheta^{(0)}\|_2^2 . 
\]

For any $j > 0$, the results holds: 
\[
\|   \bsg^{(0)}  \|_F  \leq \sqrt{T} \max_{t \in [T]} \|   g(\bx_t; \theta_0)   \|_2 \leq \calo(\sqrt{TL}),
\]
where the last inequality is held by Lemma \ref{lemma:xi1}.
Finally, given the $j>0$, 
\begin{equation} \label{eq:boundofthetak}
\|  \theta^{(j)} - \theta^{(0)} \|_2^2 \leq \sum_{i=1}^{j} \eta \left[ \bsg^{(0)} \left( [\bsg^{(0)}]^\top (\theta^{(i)}  - \theta_0) - \bsr \right)  + \lambda (\theta^{(i)} - \theta_0 )  \right]  \leq \frac{\calo(j(TL\sqrt{T/\lambda} + \sqrt{T\lambda}))}{m^{1/4}}.
\end{equation}
For (2), by standard results of gradient descent on ridge regression, $\theta^{(j)}$, and the optimum is $\btheta^{(0)} + \widehat{\btheta}_t $. Therefore, we have 
\[
\begin{aligned}
\| \theta^{(j)} - \btheta^{(0)} - \widehat{\btheta}_t \|_2^2  & \leq \left[  1 -\eta \lambda\right]^j \frac{2}{\lambda} \left(  \mathcal{L}(\btheta^{(0)}) -  \mathcal{L}(\btheta^{(0)} + \widehat{\btheta}_t) \right) \\
 \leq &  \frac{2 (1 - \eta \lambda)^j}{\lambda}  \mathcal{L}(\btheta^{(0)}) \\
= & \frac{2 (1 - \eta m \lambda)^j}{ \lambda}  \frac{\|\bsr\|^2_2}{2} \\
\leq &\frac{T(1 - \eta \lambda)^j}{ \lambda}.
\end{aligned}
\]
By setting $\lambda = 1$ and $j  = \log ((T + \calo(T^2L))^{-1})/ \log (1 - m^{- 1/4})$, we have $ \| \theta^{(j)} - \btheta_0 - \widehat{\btheta}_t \|_2^2 \leq \frac{1}{1 +  \calo(TL)} $.
Replacing $k$ and $\lambda$ in  \eqref{eq:boundofthetak} finishes the proof.
\end{proof}

\begin{lemma} \label{lemma:detazero}
Suppose $m$ satisfies the conditions in Theorem \ref{theorem:main}. With probability at least $1 - \delta$ over the initialization, the result holds:
\[
\begin{aligned}
\|  \mathbf{A}_T \|_2 &\leq 1  + \mathcal{O}(TL), \\
\log \frac{\det \mathbf{A}_T}{ \det \mathbf{I}} &\leq \widetilde{d} \log(1 + TK) + 1.
\end{aligned}
\]
\end{lemma}

\begin{proof}
Based on the Lemma \ref{lemma:xi1}, for any $t \in  [T]$, 
$ \|g(\bx_t; \btheta_0) \|_2 \leq \mathcal{O}(\sqrt{L})$.
Then, for the first item:
\[
\begin{aligned}
&\|  \mathbf{A}_T \|_2  =  \|   \mathbf{I} + \sum_{t=1}^{T} g(\bx_t; \btheta_0) g(\bx_t; \btheta_0)^\top \|_2 \\
& \leq   \| \mathbf{I} \|_2 + \| \sum_{t=1}^{T} g(\bx_t; \btheta_0) g(\bx_t; \btheta_0)^\top \|_2  \\ 
&\leq  1  +  \sum_{t=1}^{T} \| g(\bx_t; \btheta_0) \|_2^2  \leq 1  + \mathcal{O}(TL).
\end{aligned}
\]
Next, we have
\[
\log \frac{\deter(\mathbf{A}_T) }{\deter(   \mathbf{I})} = \log \deter(\mathbf{I} + \sum_{t=1}^{TK} g(\bx_t; \btheta_0) g(\bx_t; \btheta_0)^\top  ) = \deter( \mathbf{I} + \mathbf{G}_0 \mathbf{G}_0^{\top}) 
\]

Then, we have 
\[
\begin{aligned}
 &\log \det(\mathbf{I} + \mathbf{G}_0 \mathbf{G}^{\top}_0  ) \\
& = \log \deter ( \mathbf{I} + \mathbf{H}   +  (\mathbf{G}_0 \mathbf{G}^{\top}_0 - \mathbf{H})    ) \\
& \leq   \log \deter ( \mathbf{I} +  \mathbf{H}  ) + \langle ( \mathbf{I} +  \mathbf{H}   )^{-1},   (\mathbf{G}_0 \mathbf{G}^{\top}_0 - \mathbf{H})    \rangle \\
& \leq \log \deter(  \mathbf{I} +  \mathbf{H}  ) +  \| ( \mathbf{I} +  \mathbf{H}   )^{-1}    \|_{F} \|  \mathbf{G}_0 \mathbf{G}^{\top}_0 - \mathbf{H}  \|_F    \\
& \leq  \log \deter(  \mathbf{I} +  \mathbf{H}  ) +  \sqrt{T}\|  \mathbf{G}_0 \mathbf{G}^{\top}_0 - \mathbf{H}  \|_F   \\
&\leq   \log \deter(  \mathbf{I} +  \mathbf{H}  ) +  1\\
&= \widetilde{d} \log ( 1 + TK )+  1. 
\end{aligned}
\]
The first inequality is because the concavity of $\log \deter$ ; The third inequality is due to $  \| ( \mathbf{I} +  \mathbf{H} \lambda )^{-1} \|_{F} \leq  \| \mathbf{I}^{-1} \|_{F}  \leq \sqrt{T}$; The last inequality is because of the choice the $m$, based on Lemma \ref{lemma:boundgradientandNTK}; The last equality is because  of the Definition of $\hd$.
The proof is completed.
\end{proof}

\begin{lemma} \label{lemma:boundgradientandNTK}
For any $\delta \in (0, 1)$, if $m = \Omega \left( \frac{L^6 \log(TKL/\delta)}{(\epsilon/TK)^4} \right)$, then with probability at least $1 - \delta$, the results hold:
\[
\|   \mathbf{G}_0 \mathbf{G}^{\top}_0 - \mathbf{H}  \|_F \leq \epsilon.
\]
\end{lemma}
\begin{proof}
This is an application of Lemma B.1 in \cite{zhou2020neural} by properly setting $\epsilon$.
\end{proof}

\clearpage

\end{document}